\documentclass[letterpaper]{article} 
\usepackage{aaai2026}  
\usepackage{times}  
\usepackage{helvet}  
\usepackage{courier}  
\usepackage[hyphens]{url}  
\usepackage{graphicx} 
\usepackage{natbib}  
\usepackage{caption} 
\usepackage{algorithm}

\usepackage{newfloat}
\usepackage{listings}

\usepackage{comment}
\usepackage{algpseudocode}
\usepackage{multirow}    
\usepackage{amsmath}
\usepackage{subcaption}
\usepackage{enumitem}
\usepackage{amsfonts}
\usepackage{booktabs}
\usepackage{etoolbox}

\newcommand{\AAAIEquationSize}{\small} 

\AtBeginEnvironment{equation}{\AAAIEquationSize}
\AtBeginEnvironment{align}{\AAAIEquationSize}
\AtBeginEnvironment{align*}{\AAAIEquationSize}
\AtBeginEnvironment{gather}{\AAAIEquationSize}
\AtBeginEnvironment{multline}{\AAAIEquationSize}
\AtBeginEnvironment{flalign}{\AAAIEquationSize}
\AtBeginEnvironment{alignat}{\AAAIEquationSize}

\newtheorem{lemma}{Lemma}

\newtheorem{proposition}{Proposition}

\newtheorem{assumption}{Assumption}

\newcommand{\utwi}[1]{\mbox{\boldmath $ #1$}}

\newcommand{\bff}{\mathrm{\bf f}}
\newcommand{\bx}{\mathrm{\bf x}}

\newcommand{\bu}{\mathrm{\bf u}}

\newcommand{\bw}{\mathrm{\bf w}}

\newcommand{\bZ}{\mathrm{\bf Z}}
\newcommand{\bg}{\mathrm{\bf g}}

\newcommand{\bz}{\mathrm{\bf z}}

\newcommand{\bbeta}{{\utwi{\mathnormal\beta}}}
\newcommand{\bgamma}{{\utwi{\mathnormal\gamma}}}

\newcommand{\bzeta}{{\utwi{\mathnormal\zeta}}}

\newcommand{\bxi}{{\utwi{\mathnormal\xi}}}

\newcommand{\cH}{{\cal H}}

\newcommand{\cT}{{\cal T}}

\def\E{\mathbb{E}}
\def\P{\mathbb{P}}

\def\R{\mathbb{R}}

\DeclareCaptionStyle{ruled}{labelfont=normalfont,labelsep=colon,strut=off} 
\floatstyle{ruled}
\newfloat{listing}{tb}{lst}{}
\floatname{listing}{Listing}

\title{Supervised Dynamic Dimension Reduction with Deep Neural Network}

\author{
    Zhanye Luo\textsuperscript{\rm 1},
    Yuefeng Han\textsuperscript{\rm 2},
    Xiufan Yu\textsuperscript{\rm 2}
}
\affiliations{
    \textsuperscript{\rm 1}University of Chicago\\
    \textsuperscript{\rm 2}University of Notre Dame\\
    zhanye@uchicago.edu,
    yuefeng.han@nd.edu,
    xiufan.yu@nd.edu
}

\usepackage{bibentry}

\begin{document}

\maketitle

\begin{abstract}
This paper studies the problem of dimension reduction, tailored to improving time series forecasting with high-dimensional predictors. We propose a novel Supervised Deep Dynamic Principal component analysis (SDDP) framework that incorporates the target variable and lagged observations into the factor extraction process. Assisted by a temporal neural network, we construct target-aware predictors by scaling the original predictors in a supervised manner, with larger weights assigned to predictors with stronger forecasting power.
A principal component analysis is then performed on the target-aware predictors to extract the estimated SDDP factors. This supervised factor extraction not only improves predictive accuracy in the downstream forecasting task but also yields more interpretable and target-specific latent factors. Building upon SDDP, we propose a factor-augmented nonlinear dynamic forecasting model that unifies a broad family of factor-model-based forecasting approaches. To further demonstrate the broader applicability of SDDP, we extend our studies to a more challenging scenario when the predictors are only partially observable. We validate the empirical performance of the proposed method on several real-world public datasets. The results show that our algorithm achieves notable improvements in forecasting accuracy compared to state-of-the-art methods.
\end{abstract}

 \begin{links}
     \link{Code}{https://github.com/hhxsxlzy/SDDP}
 \end{links}

\section{Introduction}
Dimension reduction stands as a pivotal technique in modern data analysis to address the complexities arising from increasing data dimensionality in today's data-rich environment \citep{van2009dimensionality,sorzano2014survey}. The core idea of dimension reduction is to transform high-dimensional data into a lower-dimensional embedding while preserving essential informative features, aiming for improved model interpretability and enhanced computational efficiency. Among various dimension reduction techniques that have emerged over the decades, \emph{factor models} \citep{bai2002determining,lam2012factor} stand out as one of the most popular and widely adopted approaches. Factor models reduce dimensionality by capturing the underlying structure of high-dimensional data through a smaller set of unobserved latent variables (a.k.a., factors), which are assumed to account for the commonality among the observed variables. In supervised learning tasks with a large number of predictors, factor models are often employed as a first step to extract useful predictive information from high-dimensional predictors before applying suitable learning algorithms \citep{bair2006prediction,debashis2008,fan2017sufficient,yu2024dynamic,chen2025diffusion,luo2025factor}.

In the context of high-dimensional time series analysis, one particularly influential method is the \emph{diffusion-index forecasting model} \citep{bai2006confidence,stock2002forecasting,stock2002macroeconomic}, a factor-augmented regression that applies principal component analysis (PCA) to estimate latent factors from high-dimensional predictors, which are then used as inputs in a linear regression model to forecast the target variable. Another important direction is \emph{the sufficient forecasting approach} \cite{fan2017sufficient,yu2022nonparametric, luo2022inverse}, which uses sufficient dimension reduction (SDR) techniques \citep{li2018sufficient} to construct sufficient predictive indices for predicting the target variable. In this approach, PCA serves as an important first step to extract latent factors for subsequent estimation of the sufficient predictive indices. By contrast, classical SDR methods are designed to capture both linear and nonlinear relationships between predictors and the response variable in i.i.d. settings, as seen in foundational works such as \cite{li1991sliced, cook2002dimension, lee2014, tang2024belted}.




PCA is a well-established technique that has stood the test of time.
Despite originating decades ago, 
it continues to serve as a cornerstone of many modern methods and applications, widely used and deeply valued for its simplicity, practicality, and interpretability.
Nevertheless, traditional PCA operates under the assumption of a linear factor model and extracts components through a static, unsupervised decomposition of the data covariance matrix, which presents certain inherent limitations. 
First, \emph{the linear assumption} overlooks potential nonlinear relationships that may underlie the data structures, thereby restricting its efficacy in capturing intricate data patterns \citep{Yalcin2001NonlinearFA,scholkopf1997kernel, hinton2006reducing, gu2021autoencoder,zhou2025covariate}.
Second, \emph{the static analysis}, focusing on the cross-sectional structure of a contemporaneous panel, neglects the potential informational content embedded within the temporal dependencies of lagged observations, which can be particularly insightful in time series analysis \citep{bernanke2005measuring,ashraf2023survey,gao2024supervised}.
Third, \emph{the unsupervised nature} directs its focus solely towards maximizing variance without considering the target variable. Consequently, this may lead to the overlooking of feature directions that, despite exhibiting low variance, are highly predictive for a specific task.
These limitations motivate the development of more advanced methods and refined variants of PCA \citep{huang2022scaled,gao2024supervised}.

In this paper, we propose a Supervised Deep Dynamic PCA (SDDP) framework that efficiently constructs low-dimensional representations from high-dimensional predictors, specifically tailored for time series forecasting. Compared to traditional PCA, SDDP is nonlinear, dynamic, and supervised.
SDDP factor extraction not only improves predictive accuracy in downstream forecasting tasks but also yields more interpretable and target-specific latent factors.
Our contributions can be summarized as follows.

\begin{itemize}[leftmargin = *]
    \item SDDP explicitly incorporates the target variable and lagged observations into the training process, refining the factor extraction by aligning it more closely with the forecasting objective.
    We construct a panel of \textbf{target-aware predictors}, which scales the original predictors by their predictive power, specifically, with larger weights allocated to those predictors exhibiting stronger forecasting performance. By extracting factors in a \textbf{dynamic} and \textbf{supervised} manner, SDDP enables more effective factor extraction tailored for time series analysis. 
    \item SDDP employs advanced deep learning architectures to effectively capture the \textbf{complex nonlinear relationships} and \textbf{temporal dependencies} inherent in the data. Through a nonlinear factor model and a temporal neural network, SDDP can identify intricate nonlinear patterns that conventional linear models often overlook, contributing to more accurate and insightful predictions.
    \item Building upon SDDP, we propose \textbf{a factor-augmented nonlinear dynamic forecasting model} that \textbf{unifies a broad family of factor-model-based forecasting approaches}. By varying the underlying factor structure and selecting different link functions within the forecasting equation, the SDDP-forecasting model subsumes the classical diffusion-index model \citep{stock2002forecasting,stock2002macroeconomic} and various extensions thereof \citep{bair2006prediction,huang2022scaled,gao2024supervised}, as special cases. 
    \item Furthermore, we extend SDDP in a different direction to accommodate scenarios where the predictors are only \textbf{partially observed}. With a minor adjustment to handle missing entries, SDDP can be \textbf{adapted to covariate completion tasks} and remains effective in extracting latent factors despite the incomplete observability of the predictors.
\end{itemize}

\paragraph{Problem Setup.}
Suppose we observe a time series dataset consisting of $T$ temporal samples $\{ (\bx_t, y_t), 1\le t \le T \} $, where $\bx_t \in \mathbb{R}^{N}$ denotes the predictor vector and $y_t \in \mathbb{R}$ is the response variable. The goal is to extract supervised dynamic factors from the predictors that are most relevant for accurately forecasting the future response $y_{T+h}$.

\section{Methodology}

\subsection{Nonliner Factor Model}
For a high-dimensional observed time series predictor vector $\bx_t = (x_{1,t}, \dots, x_{N,t})^\top \in \R^N$, the objective is to forecast $y_{t+h}$ at a horizon of $h$, using the available information $\{\bx_j : j = 1, \dots, t\}$. Each component $x_{i,t}$ serves as a relevant but noisy proxy for the target, making it unlikely that a small subset alone can adequately capture the target's underlying dynamics. However, using all predictors in a conventional multivariate regression suffers from the curse of dimensionality, often leading to overfitting in-sample and poor performance out-of-sample. A common solution is to impose a factor structure on the predictors and extract a lower-dimensional set of latent factors.

We adopt a nonlinear factor model and consider a factor-augmented nonlinear dynamic forecasting model: at time $t$, the covariates $\bx_t$ and the future target $y_{t+h}$ satisfy 
\begin{align}
&x_{i,t} = x_{i,t}^\star + u_{i,t} = h_i^\star(\bff_t) + u_{i,t} = h_i^\star(\bg_t, \bzeta_t) + u_{i,t},  \label{model1} \\
&y_{t+h} = \phi( \bg_{t-q+1}, \dots, \bg_t ) + \epsilon_{t+h}, \label{model2}
\end{align}
where {\small $\bff_t=(\bg_t^\top, \bzeta_t^\top)^\top \in \R^K$} denotes the full set of latent factors. Among them, $\bg_t\in \R^{K_1}$ are the relevant factors directly related to the target $y_{t+h}$, while $\bzeta_t\in \R^{K-K_1}$ are irrelevant. The observed predictor $x_{i,t}$ consists of a common component $x_{i,t}^\star$, modeled via a possibly nonlinear loading function $h_i^\star(\cdot)$: $\mathbb{R}^K \to \mathbb{R}$ applied to the latent factors, and an idiosyncratic noise term  $u_{i,t}$. The noise vector {\small $\bu_t=(u_{1,t},...,u_{N,t})$} is assumed to be uncorrelated with the forecasting process. The function $\phi(\cdot)$ captures the nonlinear relationship between the target $y_{t+h}$ and the past $q$ lags of the relevant factors $\bg_t$. When $h_i^\star(\cdot)=b_i$ is a linear mapping, model \eqref{model1} reduces to a linear factor model, with {\small $B=(b_1^\top,...,b_p^\top)\in\R^{N\times K}$} denoting the factor loading matrix.

The factor-augmented regression model \cite{bai2006confidence,stock2002forecasting,stock2002macroeconomic} can be viewed as a special case of \eqref{model1}-\eqref{model2} under linear specifications. In their formulation, all components of $\bff_t$ are assumed to have predictive power for $y_{t+h}$. In contrast, our framework assumes that only a subset $\bg_t$ is relevant to the target, which may better reflect practical scenarios. Compared with sufficient forecasting methods \citep{fan2017sufficient,yu2022nonparametric}, model \eqref{model2} incorporates temporal dependence by allowing the response $y_{t+h}$ to depend on the past $q$ periods of the latent factors, and by permitting a nonlinear mapping $h_i^\star(\cdot)$ from the factors to the covariates.

\begin{figure*}[ht]
\centering
\includegraphics[width=0.75\linewidth]{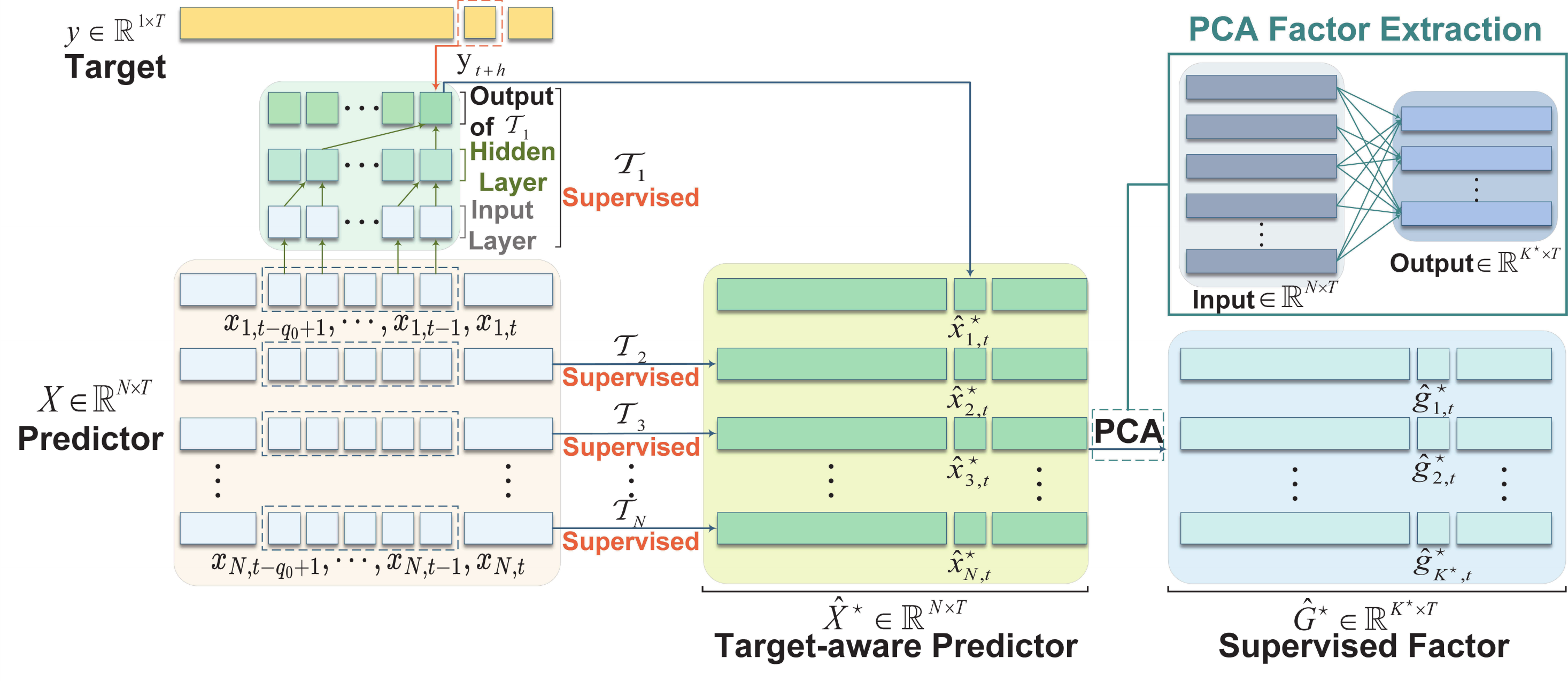}

\caption{An illustration of the proposed Supervised Deep Dynamic PCA (SDDP) algorithm. Inputs are observed predictors \(X=(\bx_1,\ldots,\bx_T)\in\mathbb{R}^{N\times T}\), target  \(y=(y_1,\ldots,y_T)^\top\in\mathbb{R}^{T}\). Output is the estimated supervised dynamic factors \(\widehat{G}^\star\).}
\label{fig:pipeline}
\end{figure*}

In the non-time-series setting with $q=1$, model \eqref{model2} has been extensively studied in the literature on sufficient dimension reduction. Building on this model, \citet{cook2002dimension,lee2014,tang2024belted} formulated the goal of sufficient mean dimension reduction as identifying a transformation $\psi: \R^{N} \to \R^{K_1}$, possibly linear or nonlinear, such that
$y_{t+h} \perp\!\!\!\perp \bx_t \mid \psi(\bx_t),$
where $\perp\!\!\!\perp$ denotes statistical independence. That is, the conditional distribution of $y_{t+h}$ given $\bx_t$ depends only on $\psi(\bx_t)$. In this sense, our model can be interpreted as a dynamic extension of sufficient dimension reduction to the time series setting.


\subsection{Supervised Deep Dynamic PCA for Time Series}

\subsubsection{Feature Reconstruction and Dimension Reduction Based on Deep Neural Network}\label{section:sdp_factors}

Given the underlying factor structure, a common approach to estimating the latent factors $\bff_t$ is to use PCA in linear settings, or autoencoders in nonlinear ones. However, under models \eqref{model1}-\eqref{model2}, both PCA and autoencoders suffer from a key limitation: they do not incorporate information from the target variable during factor extraction. Specifically, when the factors are strong, these methods cannot distinguish between target-relevant and irrelevant latent components, offering no guarantee that the top $K_1$ factors are optimal for forecasting the outcome. When the factors are weak, they may fail to separate signal from noise, resulting in biased forecasts even when using all extracted factors. To address these shortcomings, we propose a {\bf supervised deep dynamic PCA (SDDP)} approach that (i) explicitly incorporates the target variable into the factor extraction process, (ii) enables more effective dynamic ``sufficient'' dimension reduction tailored for time series prediction, and (iii) leverages flexible model architectures, using powerful neural networks to efficiently capture complex nonlinear relationships.

As illustrated in Figure~\ref{fig:pipeline}, our method consists of two key components. First, we construct a panel of {\bf target-aware predictors} $\hat x_{i,t}^\star$, where each $\hat x_{i,t}^\star$ is obtained by fitting a temporal DNN that regresses the future outcome $y_{t+h}$ on the individual predictors $\{x_{i,1},...,x_{i,t}\}$. Second, we apply conventional PCA to this panel to extract supervised factor features. This approach generalizes traditional vector-based factor models used for prediction \citep{sen2019thinkgloballyactlocally, fan2017sufficient, luo2022inverse, yu2022nonparametric,yu2024power, fan2023factor}, as it explicitly incorporates target information during feature construction. By doing so, it preserves predictive signals more effectively while simultaneously achieving dimension reduction, which substantially lowers computational cost.

In the first step, for each $i = 1, \ldots, N$, we estimate a temporal DNN that regresses the future target variable $y_{t+h}$ on the $i$-th predictor and its past lagged variables:
\begin{align}
y_{t+h} &\approx \cT_i ( x_{i,(t-q_0+1)\vee 1}, \dots, x_{i,t}; \hat\theta_i ) ,
\end{align}
where $\cT_i$ denotes a pre-specified temporal DNN architecture (e.g., Temporal Convolutional Network, TCN), $q_0$ is the window size with $q_0\ge q$, and $\hat \theta_i$ represents the learned parameters for the $i$-th neural network. The parameters $\hat\theta_i$ are obtained by minimizing the least squares loss
\begin{align*}
\hat{\theta}_i = \arg\min_{\theta_i} \sum\limits_{t} ( y_{t+h} - \mathcal{T}_i ( x_{i,(t-q_0+1)\vee 1}, \dots, x_{i,t}; \theta_i  ) )^2.
\end{align*}
After training, we use the fitted networks to construct a panel of {\bf target-aware predictors}, denoted by $\hat{\mathbf{x}}_t^\star = (\hat{x}_{1,t}^\star, \dots, \hat{x}_{N,t}^\star)^\top$, where each component is defined as
\begin{align*}
\hat{x}_{i,t}^{\star} = \mathcal{T}_i ( x_{i,(t-q_0+1)\vee 1}, \dots, x_{i,t}; \hat\theta_i  ).    
\end{align*}
This transformation embeds predictive information about the target into each feature, enhancing its relevance for downstream supervised factor extraction.

In the second step, we apply PCA to the transformed predictors $\hat \bx_t^\star$ to obtain the estimated dynamic factors $\hat \bg_t^\star$. Specifically, we compute the sample covariance matrix $\hat\Sigma=T^{-1}\sum_{t} \hat \bx_t^\star \hat\bx_t^{\star\top}\in\R^{N\times N}$, and extract the factor loading matrix $\hat B^\star\in\R^{N\times K^\star}$ as the top $K^\star$ eigenvectors\footnote{Numerous methods have been developed for determining the number of factors in factor analysis and PCA. Here, we adopt the approach by \citet{fan2022estimating} as our default choice.} of $\hat\Sigma$ scaled by $\sqrt{N}$. The estimated factors are then given by $\hat \bg_t^\star=N^{-1} \hat B^{\star\top} \hat \bx_t^\star $.
These extracted factors capture information from the original relevant latent factors $\bg_t$, as well as from their lagged values that are predictive of the future target $y_{t+h}$. Intuitively, $\hat\bg_t^\star$ serves as an estimator for the concatenated vector $(\bff_t^\top, \ldots, \bff_{t - q + 1}^\top)^\top$ in models \eqref{model1}-\eqref{model2}, under suitable conditions.
It is important to emphasize that, due to the temporal dependence structure in model \eqref{model2}, relying solely on contemporaneous predictors $x_{i,t}$ to approximate $y_{t+h}$ in the DNN fitting step 
can lead to biased parameter estimates. This, in turn, results in inefficient recovery of the relevant latent factors $\bg_t$. Incorporating lagged predictors is therefore crucial for accurate dynamic factor estimation.

A key advantage of SDDP is its ability to filter out irrelevant predictors by assigning them shrinking weights. This is critical because, unlike strong factors, weak factors often have signals that are not clearly distinguishable from noise. Without such a signal-enhancing mechanism, conventional dimension reduction methods may struggle to recover these subtle signals from the overwhelming presence of noise.

The following proposition establishes the consistency of the estimated factors in a linear setting. 

\begin{assumption}\label{asmp:linear}
Suppose a linear structure of models \eqref{model1}-\eqref{model2} 
\begin{align*}
\bx_{t} = B\bff_t + u_{t},  \quad
y_{t+h} &= \bbeta_1^\top\bg_t+\cdots+ \bbeta_q^\top\bg_{t-q+1} + \epsilon_{t+h},     
\end{align*}
where $B=(b_1^\top,...,b_p^\top)\in\R^{N\times K}$ and $1\le t\le T$. The terms $u_{t}$ and $\epsilon_{t+h}$ are i.i.d. sub-Gaussian noises with variances $\sigma_u$ and $\sigma_{\epsilon}$, respectively. We assume independence among $u_{t}$, $\epsilon_{t+h}$ and $\bff_t$. Additionally, we assume $\max_i\|b_i\|^4\le C$, and that the eigenvalues of the scaled factor loading covariance matrix $\Sigma_b=N^{-\nu} \sum_{i=1}^N b_i b_i^\top $ are bounded above and below by positive constants, where $\nu \in (0,1]$ controls factor strength, with $\nu = 1$ corresponding to strong factors and smaller $\nu$ indicating weaker factors. Finally, we assume the latent factors satisfy $\E \|\bff_t\|^4 \le C$. 
\end{assumption}

\begin{proposition}\label{prop:nonlinear}
Suppose Assumption \ref{asmp:linear} holds. Assume $q$ and $K$ are fixed. Let $\bg_t^\star=(\bff_t^\top, \ldots, \bff_{t - q + 1}^\top)^\top$. Then
there exist rotation matrices $R\in\R^{qK\times qK}$ such that
\begin{align*}
\|\hat \bg_t^\star-R\bg_t^\star\|_2=O_{\P} (N^{-\nu}+T^{-1/2} N^{-\nu/2} + N^{1-3\nu/2}T^{-3/2} ).    
\end{align*}
\end{proposition}

The well-known Wiener–Kolmogorov prediction theory \citep{kolmogorov1941interpolation, wiener1964extrapolation} is a foundational result in time series analysis. Among its key findings, it shows that under mild conditions, any weakly stationary time series can be represented as an infinite-order linear autoregressive (AR) process driven by white noise. This perspective suggests that our theoretical results, though developed in a linear framework, may be extended to more general nonlinear time series models. However, such extensions would involve substantial analytical challenges.

\begin{algorithm}[ht]
\caption{SDDP Method: Supervised Deep Dynamic PCA with Time Series Regression}
\label{alg:SDPTimesNet}
\begin{algorithmic}[1]
\Require Time series predictor $X=(x_{i,t})\in\R^{N\times T}$, target variable $\{y_t\}_{t=1}^T$, rolling window size $q_0$, temporal DNN regression model $\cH$.
\Ensure Estimated $\hat{y}_{T+h}$.

\State \textbf{Step 1: Construction of Target-aware Predictors} 
\For{each variable $i = 1, \dots, N$}
    \State Estimate parameters $\theta_i$ by solving $\hat{\theta}_i = \arg\min_{\theta_i} \sum_t ( y_{t+h} - \mathcal{T}_i ( x_{i,(t-q_0+1)\vee 1}, \dots, x_{i,t}; \theta_i ) )^2$.
    \State Construct target-aware predictors $\hat{x}_{i,t}^\star = \mathcal{T}_i ( x_{i,(t-q_0+1)\vee 1}, \dots, x_{i,t}; \hat\theta_i )$.
\EndFor

\State \textbf{Step 2: Extract Latent Factors}  
\State Form target-aware predictors $\hat\bx_t^\star=(\hat x_{1,t}^\star,...,\hat x_{N,t}^\star)^\top$ for $t=1,...,T$.
\State Compute sample covariance matrix $\hat\Sigma=T^{-1}\sum_{t} \hat \bx_t^\star \hat\bx_t^{\star\top}\in\R^{N\times N}$.
\State Determine the number of factors $K^\star$ using $\hat\Sigma$ or its normalized correlation counterpart.
\State Compute the top $K^\star$ eigenvectors $\hat U^\star$ of $\hat\Sigma$. 
\State Set the factor loading matrix $\hat B^\star=\sqrt{N} \hat U^\star$ and compute supervised factors $\hat \bg_t^\star=N^{-1}\hat B^{\star\top} \hat\bx_t^\star$.

\State \textbf{Step 3: Train Forecasting Target Model with DNN}  
\State Fit the temporal DNN model on the training data using an appropriate loss function
\begin{align*}
\hat{y}_{t+h} = \cH ( \hat\bg_{(t-q_0+1)\vee 1}^\star, \dots, \hat\bg_t^\star, y_{(t-q_0+1)\vee 1}, \dots, y_t ) .    
\end{align*}


\end{algorithmic}
\end{algorithm}

\subsubsection{Target Time Series Regression Based on Supervised Factors}\label{section:sdp_forecast}

In the previous step, we accomplished our first objective: extracting supervised dynamic factors through the SDDP method, thereby achieving supervised dynamic dimension reduction. The next step is to use these estimated latent factors $\hat \bg_t^\star$, along with their lagged values, to forecast the future outcome $y_{t+h}$. The target $y_{t+h}$ is estimated using:
\begin{align*}
\hat{y}_{t+h} = \cH ( \hat\bg_{(t-q_0+1)\vee 1}^\star, \dots, \hat\bg_t^\star, y_{(t-q_0+1)\vee 1}, \dots, y_t ),    
\end{align*}
where $\cH(\cdot)$ is a flexible nonlinear mapping capturing complex dependencies. In our empirical analysis, we explore several forecasting models for $\cH(\cdot)$, including classical deep learning architectures such as Temporal Convolutional Networks (TCN) \citep{bai2018empirical}, Long Short-Term Memory (LSTM) \citep{hochreiter1997long}, and DeepAR \citep{SALINAS20201181}, as well as recent models like DeepGLO \citep{sen2019thinkgloballyactlocally}, Autoformer \citep{wu2021autoformer}, Crossformer \citep{zhang2023crossformer}, and TimesNet \cite{wu2023timesnet}. Empirical results demonstrate that our framework consistently enhances forecasting accuracy across these models, highlighting its robustness and versatility.
Although the supervised factors from the previous subsection embed relevant temporal information and could be used without additional lags, we include both lagged factors and past outcomes in the forecasting model to further improve predictive performance. The complete algorithm is presented in Algorithm~\ref{alg:SDPTimesNet}. 
We introduced SDDP with one target variable, but it can be extended to multiple targets by aggregating their approximations in Step~1. 


\begin{table*}[ht]
\centering
\setlength{\tabcolsep}{3pt}
\small
\begin{tabular}{c*{10}{c}}
\toprule
\multirow{2}{*}{Method} 
& \multicolumn{2}{c}{Climate} 
& \multicolumn{2}{c}{Energy} 
& \multicolumn{2}{c}{FinC} 
& \multicolumn{2}{c}{Light} 
& \multicolumn{2}{c}{Weather} \\
\cmidrule(lr){2-3} \cmidrule(lr){4-5} \cmidrule(lr){6-7} \cmidrule(lr){8-9} \cmidrule(lr){10-11}
& MAE & RMSE & MAE & RMSE & MAE & RMSE & MAE & RMSE & MAE & RMSE \\
\midrule
SDDP-TCN      & \textbf{2.936} & \underline{3.673} & 41.506 & 74.279 & \underline{0.0446} & \underline{0.0564} & 2.725 & 4.965 & \textbf{2.679} & \textbf{3.369} \\
sdPCA-TCN        & 3.169 & 3.912 & 40.165 & \underline{71.045} & 0.0451 & \textbf{0.0559} & 3.000 & 4.888 & 2.704 & 3.409 \\
PCA-TCN       & 3.569 & 4.430 & 59.604 & 93.505 & 0.0484 & 0.0653 & 3.555 & 6.344 & 3.832 & 4.782 \\
Vanilla-TCN   & 3.956 & 4.838 & 62.362 & 95.822 & 0.0484 & 0.0654 & 3.494 & 6.333 & 3.991 & 4.974 \\
\midrule
SDDP-LSTM     & \underline{2.946} & \textbf{3.628} & 40.677 & 73.514 & 0.0476 & 0.0598 & 2.706 & 5.404 & \underline{2.689} & \underline{3.375} \\
sdPCA-LSTM       & 3.439 & 4.254 & 40.268 & \textbf{69.940} & 0.0476 & 0.0597 & 3.138 & 5.731 & 3.114 & 3.907 \\
PCA-LSTM      & 4.494 & 5.535 & 59.803 & 93.170 & 0.0484 & 0.0654 & 3.583 & 6.504 & 4.434 & 5.509 \\
Vanilla-LSTM  & 4.518 & 5.547 & 59.716 & 93.295 & 0.0484 & 0.0654 & 3.538 & 6.464 & 4.472 & 5.544 \\
\midrule
SDDP-DeepAR   & 3.188 & 3.953 & 52.697 & 85.275 & 0.0476 & 0.0599 & 2.912 & 5.888 & 2.954 & 3.725 \\
sdPCA-DeepAR     & 3.658 & 4.524 & 50.916 & 83.427 & 0.0536 & 0.0668 & 3.341 & 6.059 & 3.460 & 4.362 \\
PCA-DeepAR    & 4.513 & 5.570 & 59.680 & 92.882 & 0.0484 & 0.0653 & 3.629 & 6.513 & 4.445 & 5.526 \\
Vanilla-DeepAR& 4.532 & 5.575 & 58.979 & 92.850 & 0.0493 & 0.0661 & 3.548 & 6.459 & 4.487 & 5.565 \\
\midrule
SDDP-TimesNet & 4.781 & 6.057 & 39.821 & 71.133 & \textbf{0.0444} & \underline{0.0564} & 2.827 & 5.462 & 3.629 & 4.579 \\
sdPCA-TimesNet & 4.621 & 5.914 & \underline{39.094} & 71.210 & \underline{0.0446} & 0.0565 & 2.780 & 5.296 & 3.559 & 4.515 \\
PCA-TimesNet & 4.645 & 5.929 & \textbf{38.954} & 71.192 & 0.0457 & 0.0578 & \underline{2.692} & 5.343 & 3.558 & 4.509 \\
Vanilla-TimesNet & 4.753 & 6.107 & 43.980 & 73.926 & 0.0468 & 0.0592 & 2.776 & 5.389 & 3.506 & 4.424 \\
\midrule
AEAR     & 3.267 & 4.112 & 45.424 & 74.489 & 0.0469 & 0.0635 & 3.027 & 4.894 & 2.985 & 3.754 \\
DeepGLO & 3.016 & 3.881 & 43.337 & 79.443 & 0.0785 & 0.0960 & \textbf{2.025} & \textbf{4.405} & 2.943 & 3.739 \\
Autoformer & 4.024 & 4.946 & 45.123 & 74.067 & 0.0546 & 0.0687 & 2.941 & 4.815 & 3.806 & 4.795 \\
Crossformer & 4.259 & 5.445 & 46.346 & 90.817 & 0.0555 & 0.0682 & 3.110 & \underline{4.653} & 3.947 & 5.089 \\
ARIMA   & 3.315 & 4.153 & 49.166 & 81.603 & 0.0497 & 0.0666 & 3.035 & 5.387 & 3.079 & 3.816 \\
\bottomrule
\end{tabular}
\caption{Comparison of prediction accuracy (MAE and RMSE over the testing data) across various methods and datasets. Results are averaged over 100 repetitions, with the associated confidence intervals reported in Tables~S.3 in the Appendix. \textbf{Bold} indicates the best performance in each column, and \underline{underlined}  indicates the second best.}
\label{tab:Performance comparison}
\end{table*}

\subsubsection{Supervised Deep Dynamic PCA with Incomplete Data}

In real world applications, it is common to encounter missing covariate data due to situations such as sensor malfunctions, incomplete reporting, or data corruption. These missing entries pose a significant challenge for our tasks, especially when high-quality covariate information is essential for accurate dimension reduction and forecasting. To address this, we develop a dimension reduction framework tailored for scenarios with incomplete covariate data, building upon the SDDP architecture introduced in previous subsections.

In addition to the observed covariate $\bx_t$, we introduce a binary mask vector $\bw_t=(w_{1,t},...,w_{N,t})^\top \in \{0,1\}^{N}$, where

\begin{align*}
w_{i,t} =
\begin{cases}
1, & \text{if the covariate \(x_{i,t}\) is observed,} \\
0, & \text{if the covariate \(x_{i,t}\) is missing.}
\end{cases}    
\end{align*}

The first step in the SDDP framework involves supervised training of a temporal DNN for each predictor. For each $i = 1, \dots, N$, we estimate the model parameters $\theta_i$ by solving the following weighted least squares problem
\begin{align*}
\hat{\theta}_i = \arg\min_{\theta_i} \sum_{t} w_{i,t}  ( y_{t+h} - \mathcal{T}_i ( x_{i,(t-q_0+1)\vee 1}, \dots, x_{i,t}; \theta_i  )  )^2,    
\end{align*}
where $\mathcal{T}_i(\cdot; \theta_i)$ denotes the temporal DNN associated with the $i$-th predictor, and $q_0\ge q$ specifies the window size. The mask $w_{i,t}$ ensures that only observed entries are used during training.
Once trained, we use the fitted networks to construct a panel of target-aware predictors, denoted by $\hat{\mathbf{x}}_t^\star = (\hat{x}_{1,t}^\star, \dots, \hat{x}_{N,t}^\star)^\top$, where each entry is computed as
\begin{align*}
\hat{x}_{i,t}^{\star} = \mathcal{T}_i ( \tilde x_{i,(t-q_0+1)\vee 1}, \dots, \tilde x_{i,t}; \hat\theta_i ),    
\end{align*}
and each \( \tilde{x}_{i,t} \) is defined by $\tilde{x}_{i,t} =\hat{x}_{i,t}^\star$ if $ w_{i,t} = 0$, and $\tilde{x}_{i,t} =x_{i,t}$ if $ w_{i,t} = 1$.
That is, when the covariate $x_{i,t}$ is missing, we replace it with the DNN ($\cT_i$) based imputation $\hat{x}_{i,t}^\star$; otherwise, we retain the original observed value.

In the second step, mirroring previous subsection, we apply PCA to the transformed predictor panel $\hat \bx_t^\star$ to extract the estimated latent factors $\hat \bg_t^\star$. These estimated factors integrate both observed and imputed covariate information, yielding a refined representation that compensates for missing entries.

Unlike traditional matrix completion approaches that focus solely on modeling the correlation structure among covariates, our method leverages supervision from the target variable $y_{t+h}$ throughout the reconstruction process. This design aligns directly with our forecasting objective, accurately predicting $y_{t+h}$ based on the covariate history $\{\bx_1, ..., \bx_t\}$. The SDDP algorithm under covariate missingness closely follows Steps 1 and 2 of Algorithm \ref{alg:SDPTimesNet}, and is therefore omitted for brevity.

Despite its conceptual simplicity, our dynamic SDDP approach proves to be highly effective in practice. Empirical results demonstrate that it consistently outperforms baseline methods in downstream prediction tasks, especially under high levels of covariate missingness.

\begin{figure*}[t]
  \centering
  \includegraphics[height=4.8cm]{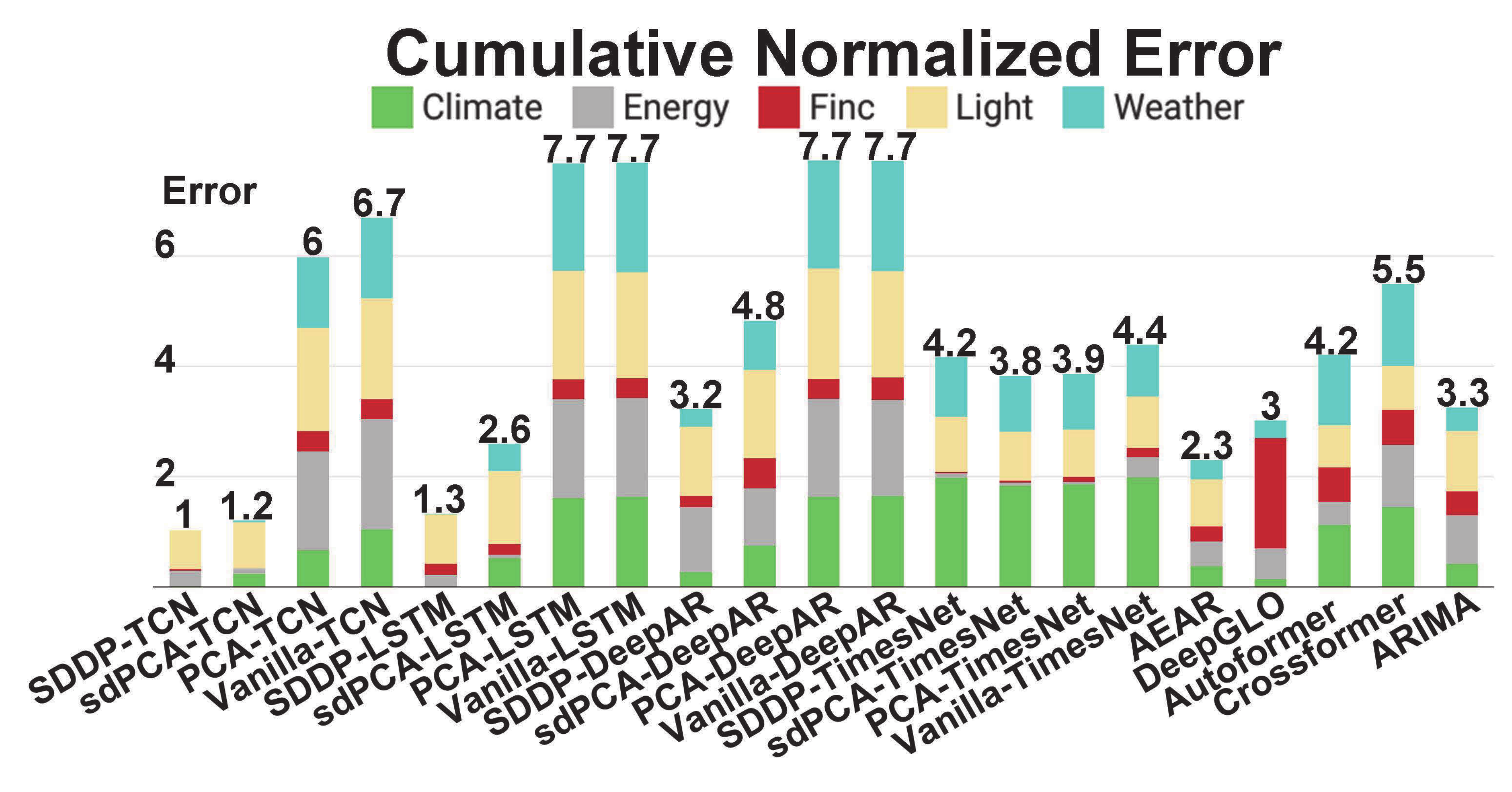}
  \hspace{0.8cm}
  \includegraphics[height=4.8cm]{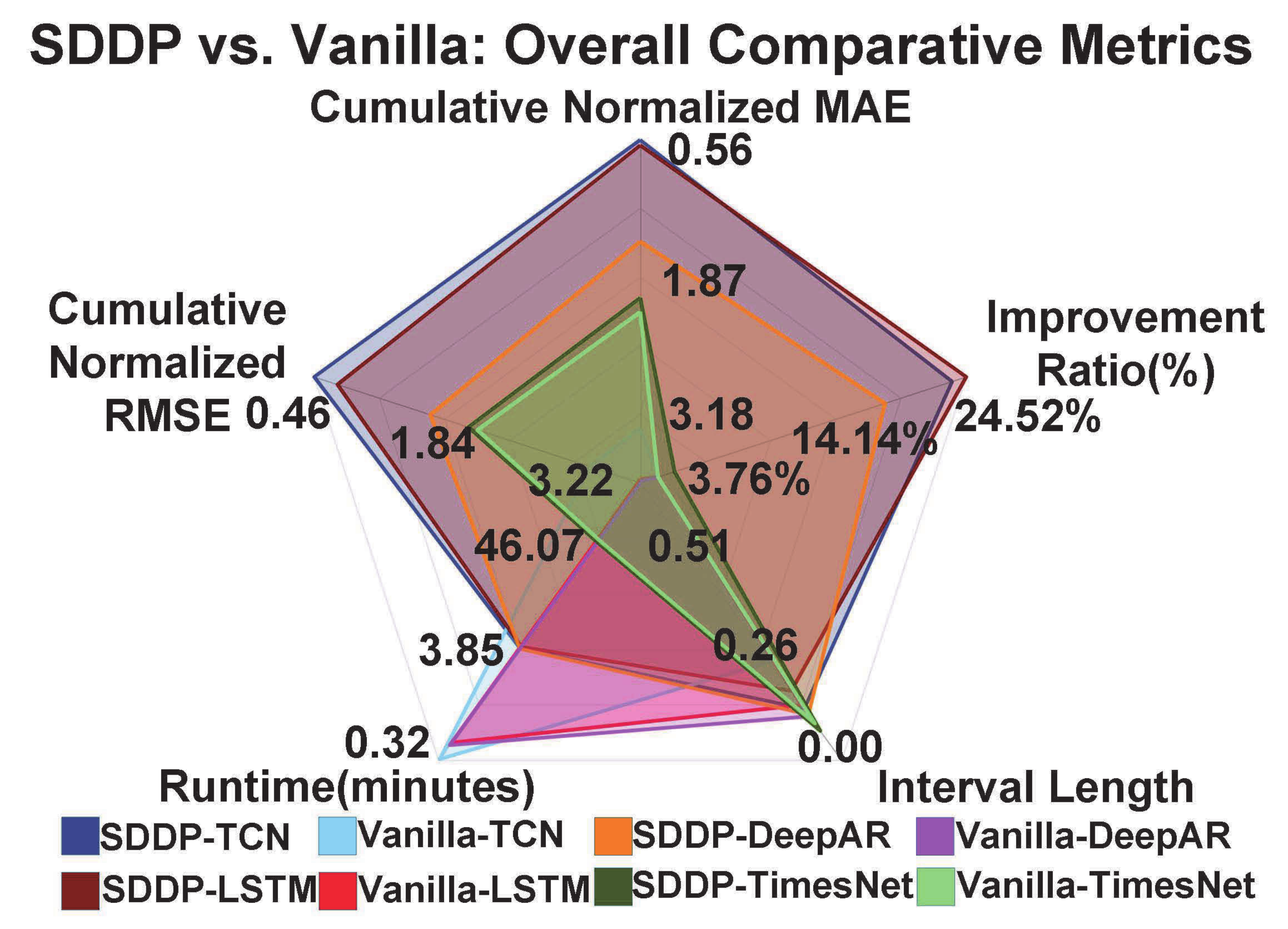}
  \caption{(a) Cumulative normalized error for each method; (b) radar chart comparing the overall performance between SDDP methods and vanilla baselines in five aspects. Detailed values and the normalization procedure are provided in the Appendix.}
  \label{fig:overall_comparison}
\end{figure*}

\section{Empirical Studies}\label{sec:empirical-results}

We demonstrate the empirical performance of the proposed SDDP across five prediction tasks using real-world high-dimensional time series datasets. Descriptions of the datasets, data sources, and preprocessing steps are provided in Table~S.1 of the Appendix. Each dataset is split chronologically with 80\% for training and 20\% for testing.

\subsubsection{Forecasting with SDDP.}
Our primary goal is to evaluate whether the proposed SDDP can extract factors that are more predictively powerful than other methods. We compare the performance of four DNN forecasting algorithms (TCN \citep{bai2018empirical}, LSTM \citep{hochreiter1997long}, DeepAR \citep{SALINAS20201181}, and TimesNet \cite{wu2023timesnet}) using factors extracted in three ways: traditional unsupervised principal component analysis (\textbf{PCA}), the supervised dynamic PCA (\textbf{sdPCA}) \cite{gao2024supervised}, and the proposed supervised deep dynamic PCA (\textbf{SDDP}). We benchmark these methods against the \textbf{Vanilla} approach that feeds high-dimensional covariates directly into DNNs without any dimension reduction. We include two factor-augmented deep learning algorithms:  Autoencoder Augmented Regression (AEAR) \citep{lei2018supervised} and DeepGLO \citep{sen2019thinkgloballyactlocally}, and two Transformer-based algorithms:  Autoformer \citep{wu2021autoformer} and Crossformer \citep{zhang2023crossformer} as well as traditional ARIMA, as additional benchmarks. 

Table \ref{tab:Performance comparison} summarizes the comparisons of prediction accuracy
using different methods.
We use the mean absolute error (MAE) and root mean squared error (RMSE) over the testing data as evaluation metrics.  Results are reported based on 100 repetitions, with the associated 95\% confidence intervals in Table~S.3 (in the Appendix). 
Since datasets vary in scale, raw MAEs and RMSEs are not directly comparable across datasets. To overcome this limitation, we provide min-max normalized MAEs and RMSEs in Table S.2, with details of the normalization procedure introduced 
in the Appendix. 
We also record the average runtime in Table~S.4 (in the Appendix) 
to further assess the practicality of various methods.

As shown in Table \ref{tab:Performance comparison}, among all the methods considered, SDDP methods consistently achieve the best performance in 4 out of 5 datasets (except for the Light data, in which DeepGLO excels). Additionally, when comparing different dimension reduction strategies within the same DNN forecasting algorithm, the proposed SDDP outperforms the sdPCA, unsupervised PCA, and Vanilla methods by a large margin in most learning algorithms (except for TimesNet, in which all dimension reduction approaches yield comparable performance). The numerical comparisons demonstrate that the dynamic, supervised, and nonlinear nature of SDDP equips it with enhanced ability to effectively capture the informative structure embedded in high-dimensional time series data, thereby boosting the model's predictive power.



\begin{table*}[t]
\centering
\small
\setlength{\tabcolsep}{4pt}
\begin{tabular}{ll*{10}{c}}
\toprule
 \multirow{3}{*}{\textbf{Dataset} }  & \multirow{3}{*}{\textbf{Method}} & \multicolumn{10}{c}{\textbf{Missing Rate}} \\ \cmidrule(lr){3-12}
& & \multicolumn{2}{c}{0\%} 
& \multicolumn{2}{c}{12.5\%} 
& \multicolumn{2}{c}{25\%} 
& \multicolumn{2}{c}{37.5\%} 
& \multicolumn{2}{c}{50\%} \\
\cmidrule(lr){3-4} \cmidrule(lr){5-6} \cmidrule(lr){7-8} \cmidrule(lr){9-10} \cmidrule(lr){11-12}
& & MAE & RMSE & MAE & RMSE & MAE & RMSE & MAE & RMSE & MAE & RMSE \\
\midrule
\multirow{9}{*}{Climate} 
& SDDP-TCN  & \textbf{2.936} & \textbf{3.673} & 2.938 & \textbf{3.645} & \textbf{2.929} & \textbf{3.642} & 2.993 & 3.693 & 3.004 & \textbf{3.703} \\
& SDDP-TimeDiff-TCN   & -- & -- & 2.992 & 3.732 & 3.066 & 3.789 & 3.006 & 3.724 & 3.055 & 3.803\\
& SDDP-MICE-TCN   & --  & --  & \textbf{2.932} & 3.667 & 2.948 & 3.688 & \textbf{2.964} & \textbf{3.691} & 2.977 & 3.708 \\
& SDDP-MissForest-TCN  & --  & --  & 2.935 & 3.676 & 2.941 & 3.689 & \textbf{2.964} & 3.702 & \textbf{2.974} & 3.710 \\
\cmidrule{2-12}
& Vanilla-TCN    & 3.956 & 4.838 & 3.979 & 4.873 & 3.946 & 4.850 & 3.948 & 4.848 & 3.812 & 4.695 \\
& Vanilla-TimeDiff-TCN & -- & -- & 3.919 & 4.811 & 3.771 & 4.686 & 3.765 & 4.717 & 3.757 & 4.689  \\
& Vanilla-MICE-TCN  & --  & --  & 4.006 & 4.897 & 3.962 & 4.845 & 3.964 & 4.844 & 3.955 & 4.834 \\
& Vanilla-MissForest-TCN  & -- & --  & 3.973 & 4.854 & 3.990 & 4.875 & 3.967 & 4.845 & 3.960 & 4.839 \\
\cmidrule{2-12}
& SDDP-TimesNet     & 4.781 & 6.057 & 4.621 & 5.899 & 4.657 & 5.953 & 4.675 & 5.951 & 4.676 & 5.981 \\
& Vanilla-TimesNet  & 4.753 & 6.107 & 4.829 & 6.204 & 4.834 & 6.201 & 4.829 & 6.192 & 4.803 & 6.146 \\
& AEAR              & 3.267 & 4.112 & 3.373 & 4.244 & 3.337 & 4.189 & 3.410 & 4.293 & 3.422 & 4.286 \\
\midrule
\multirow{9}{*}{Weather} 
& SDDP-TCN & \textbf{2.679} & \textbf{3.369} & 2.654 & 3.334 & \textbf{2.647} & 3.334 & 2.666 & \textbf{3.356} & \textbf{2.662} & \textbf{3.342} \\
& SDDP-TimeDiff-TCN   & -- & -- & \textbf{2.644} & \textbf{3.313} & 2.672 & 3.341 & 2.705 & 3.397 & 2.679 & 3.366 \\
& SDDP-MICE-TCN    & --  & --  & 2.659 & 3.346 & 2.653 & \textbf{3.327} & \textbf{2.663} & \textbf{3.356} & 2.671 & 3.356 \\
& SDDP-MissForest-TCN  & --  & --  & 2.665 & 3.348 & 2.661 & 3.331 & 2.672 & 3.359 & 2.714 & 3.406 \\
\cmidrule{2-12}
& Vanilla-TCN   & 3.991 & 4.974 & 3.954 & 4.936 & 3.887 & 4.847 & 3.923 & 4.876 & 3.940 & 4.906 \\
& Vanilla-TimeDiff-TCN & -- & -- & 3.886 & 4.864 & 3.898 & 4.875 & 3.970 & 4.938 & 3.802 & 4.772 \\
& Vanilla-MICE-TCN  & --  & --  & 3.994 & 4.979 & 3.996 & 4.987 & 3.991 & 4.969 & 4.006 & 4.995 \\
& Vanilla-MissForest-TCN  & --  & --  & 3.984 & 4.968 & 3.997 & 4.986 & 3.987 & 4.963 & 3.967 & 4.943 \\
\cmidrule{2-12}
& SDDP-TimesNet    & 3.629 & 4.579 & 3.600 & 4.570 & 3.630 & 4.611 & 3.668 & 4.665 & 3.712 & 4.731 \\
& Vanilla-TimesNet & 3.506 & 4.424 & 3.599 & 4.545 & 3.618 & 4.574 & 3.707 & 4.706 & 3.739 & 4.741 \\
& AEAR             & 2.985 & 3.754 & 3.093 & 3.890 & 3.092 & 3.872 & 3.225 & 4.055 & 3.116 & 3.897 \\
\bottomrule
\end{tabular}
\caption{Comparison of prediction accuracy  with randomly introduced covariate missingness. Results are averaged over 100 repetitions. \textbf{Bold} indicates the best performance for each metric within each dataset. ``--'' denotes fully observed data requiring no preprocessing. ``MICE'', ``MissForest'' or ``TimeDiff'' indicates the respective imputation before prediction. 
}
\label{tab:missing_pivot}
\end{table*}

We present three plots to better visualize the comparisons among various approaches. 
{Figure \ref{fig:overall_comparison}(a)} plots the cumulative normalized error across datasets to demonstrate each model's overall forecasting performance. 
The plot shows among all algorithms considered (TCN, LSTM, DeepAR, TimesNet), the SDDP-based method achieves a smaller cumulative normalized error than the corresponding sdPCA, PCA, and Vanilla methods, demonstrating evident advantages of SDDP over other dimension reduction approaches.
Figure S.1 (in the Appendix) displays the relative improvements (in percentage) of each method compared to its corresponding Vanilla baseline, revealing a consistent improvement (around 10\%--30\%) of using SDDP methods compared to Vanilla methods without performing dimension reduction on high-dimensional predictors.  
{Figure \ref{fig:overall_comparison}(b)} graphs a radar chart to provide a holistic comparison of SDDP and Vanilla methods with the four DNN forecasting algorithms in five aspects: cumulative normalized MAE, cumulative normalized RMSE, relative improvement (in percentage), runtime, and length of the confidence interval. The plot shows SDDP-TCN and SDDP-LSTM appear to achieve the overall best or near-best results in terms of predictive accuracy in the tasks in this study. The plot further emphasizes substantial advantages of SDDP over Vanilla methods in prediction accuracy, while possibly incurring slight increases in runtime. 



\subsubsection{SDDP with Partially Observed Covariates.}
Another promising application of SDDP lies in settings with partially observed covariates, specifically, when the covariate matrix contains missing values. 
Our goal is still to leverage the target time series to perform supervised dimension reduction and construct effective downstream forecasting models.
To explore this, we use the Climate and Weather datasets and introduce random missingness into the covariate matrix, with missing rates ranging from 12.5\%, 25\%, 37.5\%, 50\%. Based on the results in fully observed predictors, where TCN outperforms LSTM and DeepAR on these datasets, we focus comparison on TCN and TimesNet, using vanilla AEAR as the benchmark.
We evaluate forecasting performance using SDDP with incomplete covariates, benchmarked against the Vanilla method.
We also evaluate whether imputing missing covariates first affects subsequent factor-based prediction performance using two preprocessing techniques: {Multiple Imputation by Chained Equations} (MICE) \cite{white2011multiple} and {MissForest} \cite{stekhoven2012missforest} as well as {TimeDiff} \cite{timediff}.

Table \ref{tab:missing_pivot} summarizes the comparison of out-of-sample MAE and RMSE using different methods, based on 100 repetitions. 
Figure S.2 (in the Appendix) displays the relative improvements (in percentage) of TCN and TimesNet compared to its corresponding Vanilla baseline under partially observed covariates. 
As shown in Table~\ref{tab:missing_pivot}, forecasting with SDDP-TCN consistently outperforms other dimension reduction techniques across all levels of missingness. Figure~S.2 further illustrates that, compared to vanilla TCN without dimension reduction, SDDP-TCN yields a consistent improvement in predictive accuracy, typically around 20\%-30\%. The performance of SDDP-TimesNet is very similar to its vanilla counterpart, suggesting that transformer-based models may already capture complex structures, making additional dimension reduction less impactful. On the Climate and Weather datasets, injecting random missingness into the covariates causes only minor MAE and RMSE fluctuations, typically within 3\%, indicating that the temporal DNN component effectively imputes missing values under these conditions. 
Furthermore, SDDP-TCN with MICE, MissForest, or TimeDiff imputed data performs very similarly to original SDDP-TCN, demonstrating that the superior performance stems from the factor extraction process rather than imputation. This confirms that SDDP's performance improvement comes primarily from the framework itself rather than the imputation step.

\section{Discussion of Limitations and Extensions}\label{sec:Discussion-of-Limitations-and-Future-Extensions}

In this paper, we have introduced SDDP, a unifying framework that marries the strengths of DNNs with classical factor-model ideas to produce target-aware, low-dimensional dynamic representations for high-dimensional time series predictors. Empirical studies across five diverse datasets demonstrate that SDDP consistently outperforms unsupervised PCA, sdPCA, and a range of benchmark deep learning baselines, both in the fully observed and partially observed covariate settings.  However, SDDP does demand greater computational resources than unsupervised PCA, since training a separate deep network for each predictor can become prohibitive at large scale.

Looking ahead, the SDDP framework can be extended to handle more complex data modalities, such as tensor-valued, image, or network data, by incorporating higher-order factorization techniques (e.g., tensor decompositions) to extract supervised, dynamic features \citep{han2022rank,han2023tensor,chen2024estimation,han2024tensor,han2024simultaneous,han2024cp,kong2024teaformers,luo2025factor,zhou2025factor}. On the theoretical front, we aim to establish consistency results for nonlinear factor models, and methodologically we will investigate advanced imputation strategies (e.g., variational autoencoders) for structured missingness alongside online updating schemes to support real-time streaming applications.

\section{Acknowledgments}
We thank the anonymous reviewers for helpful suggestions. The work of Yuefeng Han is supported in part by National Science Foundation Grants DMS-2412578.  The work of Xiufan Yu is supported in part by National Institutes of Health grant R01GM152812. 


\bibliography{refs_SDDP_aaai2026}



\newpage
\appendix

\setcounter{table}{0}
\setcounter{figure}{0}
\setcounter{algorithm}{0}
\setcounter{equation}{0}
\renewcommand{\thefigure}{S.\arabic{figure}}
\renewcommand{\thetable}{S.\arabic{table}}
\renewcommand{\thealgorithm}{S.\arabic{algorithm}}
\renewcommand{\theequation}{S.\arabic{equation}}

\section*{\Large Appendices}

\section{Supplement to Empirical Studies} 

\subsection{Dataset Information} \label{append:data-info}

High-dimensional time series data are widely present in various real-world domains, such as finance, meteorology, and energy. 
We evaluate the empirical performance of the proposed SDDP across five prediction tasks using real-world high-dimensional time series datasets. A summary of the data sources and preprocessing steps is provided in Table~\ref{tab:datasets}.

\begin{figure*}[h]
  \centering
   \begin{minipage}{0.49\textwidth}
   \centering
    \includegraphics[width=0.9\textwidth]{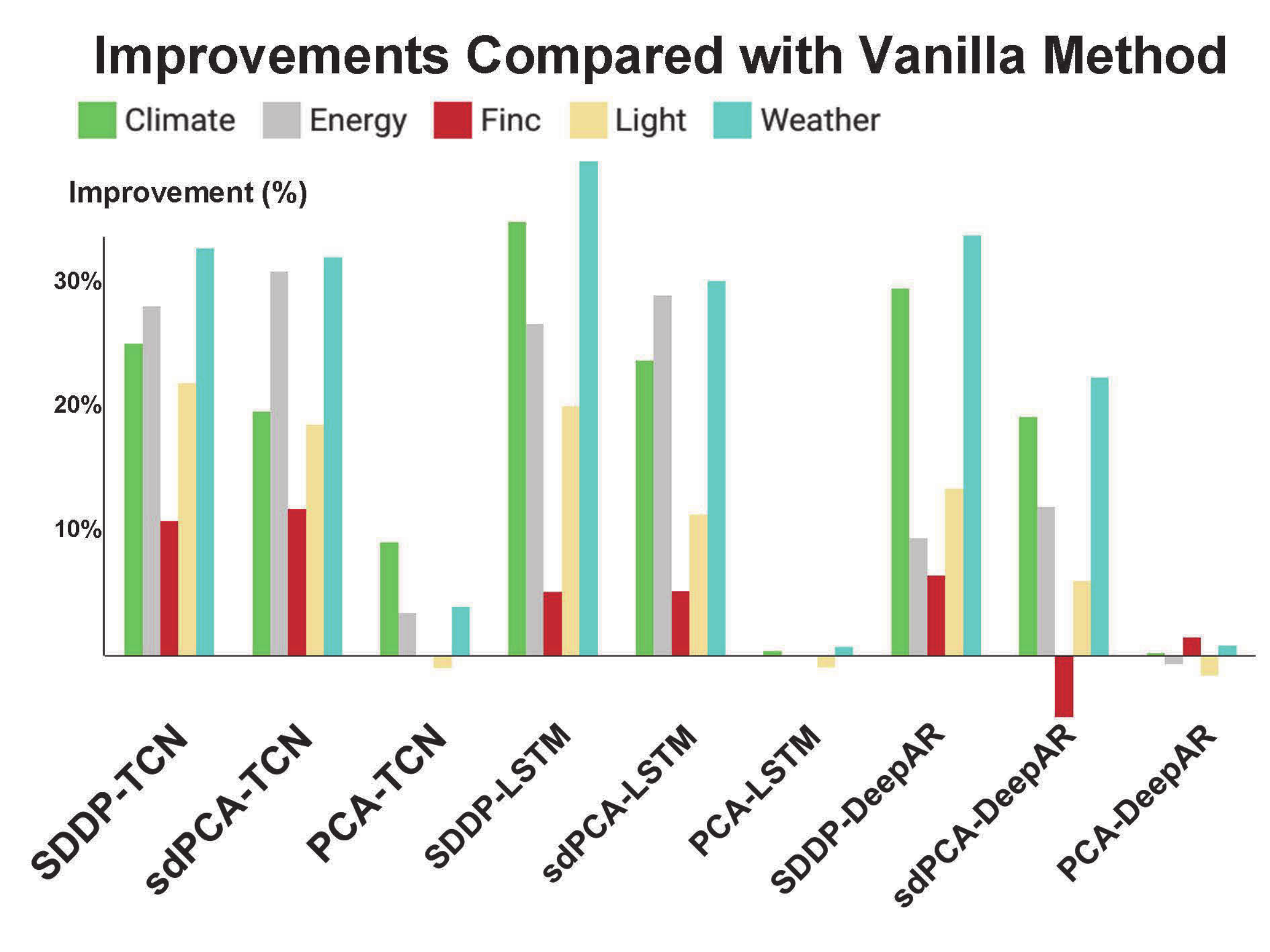}
    \captionof{figure}{\small Relative improvements (in percentage) of SDDP, sdPCA, and PCA over the Vanilla baseline.}
    \label{fig:improvement}
  \end{minipage}
  \hfill 
    \begin{minipage}{0.49\textwidth}
   \centering
    \includegraphics[width=0.9\textwidth]{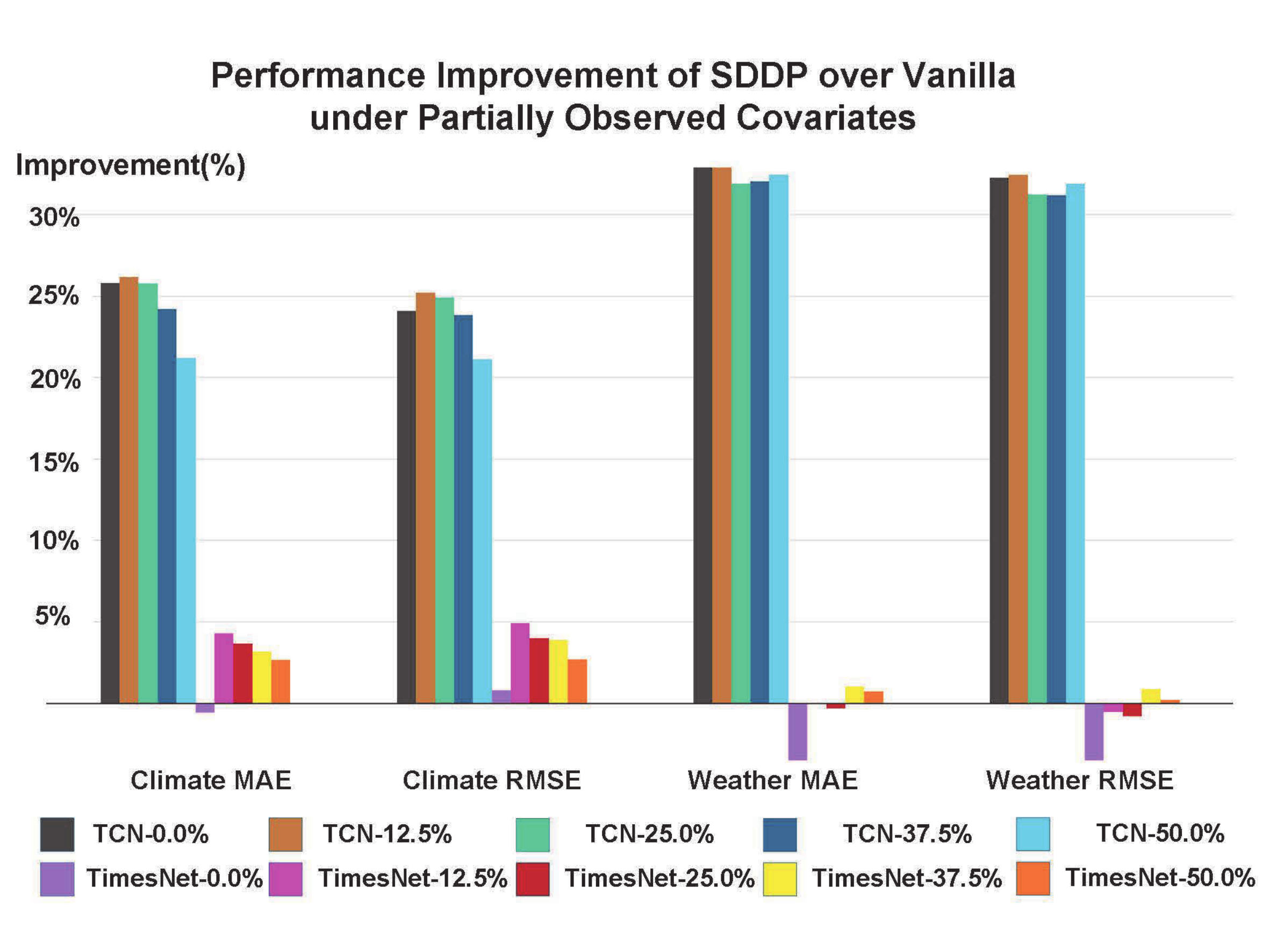}
    \captionof{figure}{\small Relative improvements (in percentage) of SDDP vs. Vanilla baselines in the presence of missing values.}
    \label{fig:sdp_vs_vanilla}
  \end{minipage}
\end{figure*}

\begin{table*}[ht]
\centering
\small
\begin{tabular}{llll}
\toprule
Full Name                         & Alias        & Source                                   & Preprocessing \\ 
\midrule
Characteristic Signals            & FinC         & Kozak \cite{Kozak2019KernelTF}           & long-short portfolio construction \\
Jena Climate                      & Climate      & Max Planck Institute  & Daily averaging \\
TimeSeries Weather Dataset        & Weather      & Kaggle                                   & Daily averaging \\
Appliances Energy Prediction      & Energy & UCI \cite{Candanedo2017DataDP}            & 30-min averaging \\
& Light & UCI \cite{Candanedo2017DataDP}            & 30-min averaging \\
\bottomrule
\end{tabular}
\caption{Detailed information of the high-dimensional time series datasets.}
\label{tab:datasets}
\end{table*}

In the financial domain, we utilize the ``Characteristic Signals'' dataset collected by Kozak \cite{Kozak2019KernelTF}, covering the period from July 1963 to December 2019, which consists of 50 signals measuring stock returns. During preprocessing, we augment the dataset by adding the return of each stock. Subsequently, we construct a long-short portfolio using the first and last 10\% of stocks based on these signals. The objective of this task is to predict the portfolio return using characteristic signals as covariates. In the subsequent empirical study, we refer to this dataset as \texttt{FinC}. 

In the meteorology domain, we evaluate our model on two datasets. The first is the ``Jena Climate'' dataset, a weather time series recorded at the Weather Station of the Max Planck Institute for Biogeochemistry in Jena, Germany, spanning from 2009 to 2016. The second dataset, ``TimeSeries Weather Dataset'', is sourced from Kaggle. To ensure consistency in granularity and reduce computational costs during model training, we preprocess both datasets by computing their daily averages. In the subsequent empirical study, we refer to these two datasets as \texttt{Climate} and \texttt{Weather}, respectively.  

In the energy domain, we utilize the ``Appliances Energy Prediction'' dataset from the UCI repository\cite{Candanedo2017DataDP}. Given the high-frequency fluctuations in the data, we apply a half-hour averaging to smooth the time series. Regarding the selection of target variables, we consider two aspects: the total energy consumption of the building (Energy) and the energy consumption of light fixtures (Light). In the following empirical analysis, we refer to these two prediction tasks as \texttt{Energy} and \texttt{Light}, respectively.

\subsection{Cumulative Normalized Error} \label{append:cumulative-error}

Since datasets vary in scale, raw MAEs and RMSEs are not directly comparable across datasets. To overcome this limitation, we apply Min-Max normalization to the prediction errors (MAE and RMSE) reported in Table \ref{tab:Performance comparison}.



The Cumulative Normalized Error is defined to enable fair comparison across datasets with different scales. Specifically, for each dataset, we apply Min-Max normalization to the prediction errors of all models. Let \( e_{i,d} \) denote the error (either MAE or RMSE) of model \( i \) on dataset \( d \), then the normalized error \( \tilde{e}_{i,d} \) is computed as:
\[
\tilde{e}_{i,d} = \frac{e_{i,d} - \min_j e_{j,d}}{\max_j e_{j,d} - \min_j e_{j,d}}.
\]

\begin{table*}[htbp]
\centering
\small
\setlength{\tabcolsep}{4pt}
\renewcommand{\arraystretch}{0.95}
\begin{tabular}{l
  *{2}{c}
  *{2}{c}
  *{2}{c}
  *{2}{c}
  *{2}{c}
}
\toprule
\textbf{Method} &
\multicolumn{2}{c}{\textbf{Climate}} &
\multicolumn{2}{c}{\textbf{Energy}} &
\multicolumn{2}{c}{\textbf{Finc}} &
\multicolumn{2}{c}{\textbf{Light}} &
\multicolumn{2}{c}{\textbf{Weather}} \\
 & MAE & RMSE & MAE & RMSE & MAE & RMSE & MAE & RMSE & MAE & RMSE \\
\midrule
SDDP-TCN         & 0.0000 & 0.0180 & 0.1090 & 0.1676 & 0.0128 & 0.0177 & 0.4363 & 0.2657 & 0.0000 & 0.0000 \\
sdPCA-TCN        & 0.1263 & 0.1147 & 0.0517 & 0.0427 & 0.0000 & 0.0000 & 0.6075 & 0.2294 & 0.0139 & 0.0181 \\
PCA-TCN          & 0.3433 & 0.3234 & 0.8822 & 0.9105 & 0.1223 & 0.2381 & 0.9534 & 0.9196 & 0.6376 & 0.6434 \\
Vanilla-TCN      & 0.5531 & 0.4881 & 1.0000 & 1.0000 & 0.1240 & 0.2389 & 0.9158 & 0.9143 & 0.7256 & 0.7310 \\
\midrule
SDDP-LSTM        & 0.0056 & 0.0000 & 0.0736 & 0.1381 & 0.1002 & 0.0994 & 0.4244 & 0.4739 & 0.0054 & 0.0027 \\
sdPCA-LSTM       & 0.2729 & 0.2523 & 0.0561 & 0.0000 & 0.0991 & 0.0986 & 0.6938 & 0.6291 & 0.2404 & 0.2450 \\
PCA-LSTM         & 0.8446 & 0.7692 & 0.8907 & 0.8975 & 0.1233 & 0.2382 & 0.9708 & 0.9955 & 0.9705 & 0.9746 \\
Vanilla-LSTM     & 0.8577 & 0.7742 & 0.8870 & 0.9024 & 0.1243 & 0.2386 & 0.9433 & 0.9766 & 0.9919 & 0.9904 \\
\midrule
SDDP-DeepAR      & 0.1371 & 0.1311 & 0.5871 & 0.5925 & 0.1002 & 0.1024 & 0.5529 & 0.7036 & 0.1519 & 0.1620 \\
sdPCA-DeepAR     & 0.3913 & 0.3615 & 0.5110 & 0.5211 & 0.2740 & 0.2740 & 0.8205 & 0.7844 & 0.4320 & 0.4521 \\
PCA-DeepAR       & 0.8547 & 0.7832 & 0.8854 & 0.8864 & 0.1225 & 0.2381 & 1.0000 & 1.0000 & 0.9769 & 0.9824 \\
Vanilla-DeepAR   & 0.8650 & 0.7854 & 0.8555 & 0.8852 & 0.1492 & 0.2569 & 0.9493 & 0.9744 & 1.0000 & 1.0000 \\
\midrule
SDDP-TimesNet    & 1.0000 & 0.9797 & 0.0370 & 0.0461 & 0.0073 & 0.0159 & 0.4996 & 0.5016 & 0.5257 & 0.5510 \\
sdPCA-TimesNet   & 0.9135 & 0.9220 & 0.0060 & 0.0491 & 0.0125 & 0.0183 & 0.4705 & 0.4227 & 0.4867 & 0.5218 \\
PCA-TimesNet     & 0.9263 & 0.9281 & 0.0000 & 0.0484 & 0.0433 & 0.0495 & 0.4159 & 0.4448 & 0.4863 & 0.5192 \\
Vanilla-TimesNet & 0.9851 & 1.0000 & 0.2147 & 0.1540 & 0.0767 & 0.0858 & 0.4683 & 0.4667 & 0.4575 & 0.4804 \\
\midrule
AEAR             & 0.1798 & 0.1951 & 0.2764 & 0.1757 & 0.0801 & 0.1914 & 0.6243 & 0.2321 & 0.1694 & 0.1753 \\
DeepGLO          & 0.0433 & 0.1021 & 0.1872 & 0.3672 & 1.0000 & 1.0000 & 0.0000 & 0.0000 & 0.1459 & 0.1683 \\
Autoformer & 0.5899 & 0.5315 & 0.2635 & 0.1595 & 0.3045 & 0.3208 & 0.5707 & 0.1946 & 0.6231 & 0.6494 \\
Crossformer & 0.7174 & 0.7330 & 0.3158 & 0.8066 & 0.3293 & 0.3091 & 0.6764 & 0.1176 & 0.7011 & 0.7831 \\
ARIMA            & 0.2059 & 0.2115 & 0.4362 & 0.4506 & 0.1596 & 0.2700 & 0.6297 & 0.4658 & 0.2212 & 0.2036 \\
\bottomrule
\end{tabular}
\caption{Normalized MAE and RMSE across datasets. For each dataset and metric, model performances are min-max normalized such that the best model maps to 0 and the worst to 1. These values reflect relative error comparisons among models and do not imply absence of prediction error.}\label{tab:cumulative-normalized-error}
\end{table*}

To compute the Cumulative Normalized Error for each model, we sum its normalized MAE and RMSE across all datasets. Formally, for a given model \( i \), the cumulative error is calculated as:
\[
\text{NCE}_i = \sum_{d \in \mathcal{D}} \left( \tilde{e}^{\text{MAE}}_{i,d} + \tilde{e}^{\text{RMSE}}_{i,d} \right),
\]
where \( \tilde{e}^{\text{MAE}}_{i,d} \) and \( \tilde{e}^{\text{RMSE}}_{i,d} \) denote the Min-Max normalized MAE and RMSE of model \( i \) on dataset \( d \), and \( \mathcal{D} \) represents the set of all datasets. This cumulative score provides an aggregate measure of model performance, with lower values indicating better overall accuracy.

The min-max normalized MAEs and RMSEs are summarized in Table \ref{tab:cumulative-normalized-error}. 
After normalization, we sum the normalized errors across all datasets for each model to obtain its cumulative normalized error. The cumulative normalized errors across datasets are plotted in Figure {\ref{fig:overall_comparison}(a)} to demonstrate each model's overall forecasting performance. A smaller cumulative error indicates better overall performance. 


\subsection{Supplemental Numerical Results}
Figure \ref{fig:improvement} displays the relative improvements (in percentage) of each method compared to its corresponding Vanilla baseline, revealing a consistent improvement (around 10\%--30\%) of using SDDP-based methods compared to Vanilla methods without performing dimension reduction on high-dimensional predictors. 

Figure~\ref{fig:sdp_vs_vanilla} displays the relative improvements (in percentage) of TCN and TimesNet compared to its corresponding Vanilla baseline under partially observed covariates. The plot illustrates that, in settings with partially observed covariates, SDDP-TCN yields a consistent improvement in predictive accuracy, typically around 20\%-30\%, compared to vanilla TCN without dimension reduction.

\subsection{Confidence Interval} \label{append:confidence-intervals}

We also quantified the uncertainty associated with each model's forecasting performance by constructing 95\% confidence intervals. This was achieved by independently running each model 100 times and computing the empirical confidence intervals for both MAE and RMSE across all datasets. The resulting intervals are summarized in Tables~\ref{tab:Confidence interval details}, providing insight into the stability and robustness of the models under repeated trials.

\begin{table*}[htbp]
\centering

\setlength{\tabcolsep}{3pt}
\renewcommand{\arraystretch}{0.95}
\resizebox{\textwidth}{!}{%
\begin{tabular}{l*{5}{cc}}
\toprule
\multirow{2}{*}{\textbf{Method}} 
& \multicolumn{2}{c}{\textbf{Climate}} 
& \multicolumn{2}{c}{\textbf{Energy}} 
& \multicolumn{2}{c}{\textbf{FinC}} 
& \multicolumn{2}{c}{\textbf{Light}} 
& \multicolumn{2}{c}{\textbf{Weather}} \\
\cmidrule(lr){2-3} \cmidrule(lr){4-5} \cmidrule(lr){6-7} \cmidrule(lr){8-9} \cmidrule(lr){10-11}
& MAE & RMSE & MAE & RMSE & MAE & RMSE & MAE & RMSE & MAE & RMSE \\
\midrule
SDDP-TCN & [2.929, 2.942] & [3.665, 3.680] & [42.026, 42.855] & [74.803, 75.046] & [0.044, 0.045] & [0.056, 0.057] & [2.696, 2.754] & [4.959, 4.971] & [2.675, 2.683] & [3.364, 3.374] \\
sdPCA-TCN & [3.153, 3.184] & [3.893, 3.932] & [39.874, 40.455] & [71.006, 71.084] & [0.045, 0.045] & [0.056, 0.056] & [2.970, 3.029] & [4.882, 4.895] & [2.695, 2.713] & [3.395, 3.422] \\
PCA-TCN & [3.515, 3.623] & [4.367, 4.493] & [59.304, 59.903] & [93.122, 93.889] & [0.048, 0.048] & [0.065, 0.065] & [3.529, 3.580] & [6.321, 6.366] & [3.792, 3.871] & [4.734, 4.830] \\
Vanilla-TCN & [3.922, 3.990] & [4.798, 4.878] & [61.942, 62.782] & [95.280, 96.364] & [0.048, 0.048] & [0.065, 0.065] & [3.469, 3.519] & [6.303, 6.362] & [3.970, 4.012] & [4.950, 4.999] \\
\midrule
SDDP-LSTM & [2.926, 2.966] & [3.602, 3.654] & [41.743, 42.470] & [74.199, 74.702] & [0.047, 0.048] & [0.059, 0.060] & [2.679, 2.763] & [5.517, 5.636] & [2.672, 2.705] & [3.354, 3.396] \\
sdPCA-LSTM & [3.395, 3.483] & [4.198, 4.309] & [40.045, 40.490] & [69.734, 70.147] & [0.047, 0.048] & [0.059, 0.060] & [3.078, 3.198] & [5.661, 5.801] & [3.080, 3.147] & [3.866, 3.949] \\
PCA-LSTM & [4.511, 4.525] & [5.540, 5.555] & [59.406, 60.025] & [93.022, 93.568] & [0.048, 0.048] & [0.065, 0.065] & [3.526, 3.551] & [6.446, 6.481] & [4.467, 4.478] & [5.538, 5.551] \\
Vanilla-LSTM & [4.488, 4.499] & [5.529, 5.541] & [59.611, 59.995] & [93.004, 93.337] & [0.048, 0.048] & [0.065, 0.065] & [3.572, 3.593] & [6.489, 6.518] & [4.429, 4.439] & [5.504, 5.515] \\
\midrule
SDDP-DeepAR & [3.155, 3.222] & [3.912, 3.994] & [52.486, 52.907] & [85.213, 85.337] & [0.047, 0.048] & [0.059, 0.061] & [2.857, 2.967] & [5.821, 5.955] & [3.012, 3.065] & [3.800, 3.864] \\
sdPCA-DeepAR & [3.583, 3.732] & [4.430, 4.619] & [50.044, 51.788] & [83.028, 83.827] & [0.052, 0.055] & [0.065, 0.068] & [3.268, 3.415] & [5.964, 6.153] & [3.414, 3.506] & [4.306, 4.418] \\
PCA-DeepAR & [4.508, 4.517] & [5.565, 5.575] & [59.419, 59.941] & [92.637, 93.128] & [0.048, 0.048] & [0.065, 0.065] & [3.618, 3.640] & [6.501, 6.525] & [4.441, 4.450] & [5.522, 5.531] \\
Vanilla-DeepAR & [4.526, 4.537] & [5.569, 5.582] & [58.737, 59.221] & [92.626, 93.074] & [0.049, 0.049] & [0.066, 0.066] & [3.537, 3.559] & [6.448, 6.471] & [4.483, 4.491] & [5.561, 5.570] \\
\midrule
SDDP-TimesNet & [4.739, 4.822] & [5.992, 6.122] & [39.737, 39.906] & [71.049, 71.217] & [0.044, 0.045] & [0.056, 0.057] & [2.814, 2.839] & [5.454, 5.471] & [3.611, 3.648] & [4.555, 4.603] \\
sdPCA-TimesNet & [4.576, 4.665] & [5.846, 5.982] & [38.991, 39.196] & [70.121, 70.300] & [0.044, 0.045] & [0.056, 0.057] & [2.769, 2.792] & [5.292, 5.300] & [3.547, 3.571] & [4.500, 4.531] \\
PCA-TimesNet & [4.608, 4.681] & [5.873, 5.985] & [38.869, 39.040] & [71.126, 71.258] & [0.045, 0.046] & [0.057, 0.058] & [2.686, 2.699] & [5.337, 5.348] & [3.541, 3.575] & [4.486, 4.533] \\
Vanilla-TimesNet & [4.715, 4.792] & [6.046, 6.168] & [43.811, 44.150] & [73.823, 74.028] & [0.045, 0.046] & [0.057, 0.058] & [2.769, 2.784] & [5.382, 5.395] & [3.490, 3.523] & [4.402, 4.446] \\
\midrule
AEAR & [3.198, 3.336] & [4.027, 4.197] & [44.787, 46.061] & [73.701, 75.277] & [0.047, 0.047] & [0.063, 0.064] & [2.941, 3.113] & [4.856, 4.932] & [2.909, 3.062] & [3.663, 3.844] \\
Autoformer &
[3.949, 4.098] & [4.861, 5.031] & [44.631, 45.615] & [73.532, 74.603] &
[0.533, 0.560] & [0.671, 0.703] & [2.917, 2.964] &
[4.788, 4.842] & [3.762, 3.849] & [4.740, 4.850] \\
Crossformer &
[4.229, 4.290] & [5.409, 5.482] & [46.204, 46.350] & [90.721, 90.819] &
[0.051, 0.060] & [0.064, 0.073] & [3.083, 3.137] &
[4.628, 4.677] & [3.914, 3.980] & [5.048, 5.130] \\
DeepGLO & [2.786, 3.245] & [3.611, 4.152] & [42.085, 44.589] & [78.392, 80.494] & [0.071, 0.086] & [0.089, 0.103] & [2.006, 2.045] & [4.394, 4.416] & [2.783, 3.103] & [3.548, 3.929] \\
ARIMA & [3.315, 3.315] & [4.153, 4.153] & [49.166, 49.166] & [81.603, 81.603] & [0.050, 0.050] & [0.067, 0.067] & [3.035, 3.035] & [5.387, 5.387] & [3.079, 3.079] & [3.816, 3.816] \\
\bottomrule
\end{tabular}%
}
\caption{95\% Confidence Intervals for MAE and RMSE across datasets and models. Each model and method combination is run 100 times to compute the confidence intervals.}
\label{tab:Confidence interval details}
\end{table*}

\subsection{Run Time} \label{append:runtime}

To further assess the practicality and robustness of the proposed approach, we additionally recorded the runtime of each algorithm under a unified computing environment. Specifically, all models were executed on HPE ProLiant DL385 Gen10 servers equipped with dual 24-core AMD EPYC 7451 CPUs @ 2.30GHz, 128 GB RAM, and 2 TB SSD storage. To ensure a fair comparison, early stopping was enabled with a patience of 3 across all experiments. The corresponding single-metric runtime statistics for each model and dataset are reported in Table~\ref{tab:Runtime}.

\begin{table*}[htbp]
\centering
\small
\setlength{\tabcolsep}{4pt}
\renewcommand{\arraystretch}{1.0}

\begin{tabular}{l|ccccc}
\toprule
\textbf{Method} & \textbf{Climate} & \textbf{Energy} & \textbf{FinC} & \textbf{Light} & \textbf{Weather} \\
\midrule
SDDP-TCN        & 1.262  & 8.370   & 0.120  & 6.4427  & 3.514 \\
sdPCA-TCN       & 1.326   & 4.875   & 0.280  & 4.167   & 3.329 \\
PCA-TCN         & 0.308   & 0.350   & 0.122  & 0.337   & 0.492 \\
Vanilla-TCN     & 0.365   & 0.415   & 0.130  & 0.230   & 0.520 \\
\midrule
SDDP-LSTM       & 1.752   & 6.976   & 0.485  & 6.537   & 4.708 \\
sdPCA-LSTM      & 1.449   & 4.762   & 0.535  & 4.768   & 4.242 \\
PCA-LSTM        & 0.398   & 0.828   & 0.150  & 0.317   & 0.683 \\
Vanilla-LSTM    & 0.337   & 0.888   & 0.156  & 0.317   & 0.697 \\
\midrule
SDDP-DeepAR     & 1.694   & 6.131   & 0.4888 & 5.995   & 5.257 \\
sdPCA-DeepAR    & 1.302   & 4.160   & 0.224  & 4.536   & 3.835 \\
PCA-DeepAR      & 0.387   & 0.690   & 0.180  & 0.313   & 0.603 \\
Vanilla-DeepAR  & 0.398   & 0.750   & 0.188  & 0.325   & 0.601 \\
\midrule
SDDP-TimesNet   & 26.753  & 56.102  & 4.910  & 42.100  & 82.683 \\
sdPCA-TimesNet  & 25.665  & 56.150  & 4.393  & 47.140  & 84.344 \\
PCA-TimesNet    & 26.967  & 62.995  & 4.540  & 56.165  & 90.753 \\
Vanilla-TimesNet& 29.582  & 52.618  & 3.977  & 55.405  & 81.396 \\
\midrule
AEAR            & 0.212   & 0.420   & 0.155  & 0.418   & 0.472 \\
DeepGLO         & 114.820 & 203.758 & 24.158 & 221.771 & 232.329 \\
Autoformer      & 4.383   & 32.808  & 3.483  & 30.975  & 32.980 \\
Crossformer     & 3.533   & 33.306  & 3.436  & 27.530  & 34.210 \\
ARIMA           & 1.583   & 4.733   & 0.0933 & 4.350   & 4.117 \\
\bottomrule
\end{tabular}
\caption{Comparisons of runtime across various methods and datasets. Each model is trained with early stopping (patience = 3) on a 24-core AMD EPYC 7451 CPU @ 2.30GHz and 128 GB RAM. }

\label{tab:Runtime}
\end{table*}

\subsection{Hyperparameters and Sensitive Analysis \label{append:sensitive_analysis}}

By default, two key hyperparameters are set as follows: the number of factors $K^*$ (retained during PCA-based dimension reduction) is determined using the eigenvalue criterion of the correlation matrix \citep{fan2022estimating}, retaining only components with eigenvalues greater than one, with a maximum of seven factors; and the window length $q$ (the temporal window for time series modeling) is set to $\min\{200,T/20\}$.

We conducted a sensitivity analysis on $K^*$ and $q$. The $q$-length experiments exclude the \textit{Finc} dataset because its short time horizon ($T/20 < 50$) prevents fair comparison under longer temporal windows.

As shown in Table~\ref{tab:Sensitive_K}, $K^*$ has limited influence on model performance, typically within 3\%. Excluding the \textit{Finc} dataset--where the effect is around 10\% due to high dimensionality ($p = 50$) and noisy financial data--the sensitivity remains small. The window length $q$ has minimal effect (Table~\ref{tab:Sensitive_p}), generally within 2\%.

\begin{table*}[htbp]
\centering
\small
\setlength{\tabcolsep}{4pt}
\renewcommand{\arraystretch}{0.9}
\begin{tabular}{l|c|cc|cc|cc|cc|cc}
\toprule
\multicolumn{2}{c}{} & \multicolumn{10}{c}{\textbf{Dataset}} \\
\cmidrule(lr){3-12}
\multicolumn{1}{c}{\textbf{Model}} & \multicolumn{1}{c}{\textbf{N\_Components ($K$)}} &
\multicolumn{2}{c}{Climate} &
\multicolumn{2}{c}{Energy} &
\multicolumn{2}{c}{FinC} &
\multicolumn{2}{c}{Light} &
\multicolumn{2}{c}{Weather} \\
\cmidrule(lr){3-4}
\cmidrule(lr){5-6}
\cmidrule(lr){7-8}
\cmidrule(lr){9-10}
\cmidrule(lr){11-12}
& &
MAE & RMSE &
MAE & RMSE &
MAE & RMSE &
MAE & RMSE &
MAE & RMSE \\
\midrule
\multirow{5}{*}{SDDP-TCN}
 & 2  & 2.966 & 3.709 & 41.506 & 74.279 & 0.0453 & 0.0569 & 2.791 & 4.942 & 2.690 & 3.386 \\
 & 3  & 2.941 & 3.677 & 41.340 & 74.668 & 0.0457 & 0.0575 & 2.802 & 4.933 & 2.673 & 3.363 \\
 & 5  & 2.925 & 3.661 & 42.487 & 74.967 & 0.0471 & 0.0591 & 2.727 & 4.966 & 2.631 & 3.316 \\
 & 8  & 2.888 & 3.610 & 42.267 & 74.880 & 0.0487 & 0.0609 & 2.757 & 4.957 & 2.629 & 3.320 \\
 & 10 & 2.883 & 3.586 & 42.601 & 75.816 & 0.0501 & 0.0626 & 2.713 & 4.959 & 2.640 & 3.335 \\
\midrule
\multirow{5}{*}{SDDP-LSTM}
 & 2  & 2.969 & 3.642 & 40.677 & 73.514 & 0.0458 & 0.0575 & 2.745 & 5.308 & 2.671 & 3.347 \\
 & 3  & 2.934 & 3.605 & 41.360 & 74.268 & 0.0465 & 0.0585 & 2.706 & 5.404 & 2.678 & 3.363 \\
 & 5  & 2.957 & 3.651 & 41.123 & 74.786 & 0.0475 & 0.0597 & 2.712 & 5.513 & 2.717 & 3.421 \\
 & 8  & 2.998 & 3.716 & 41.398 & 74.635 & 0.0488 & 0.0607 & 2.715 & 5.597 & 2.807 & 3.549 \\
 & 10 & 2.955 & 3.673 & 42.632 & 74.894 & 0.0499 & 0.0621 & 2.702 & 5.549 & 2.887 & 3.660 \\
\midrule
\multirow{5}{*}{SDDP-DeepAR}
 & 2  & 3.339 & 4.149 & 52.695 & 85.274 & 0.0457 & 0.0569 & 3.050 & 5.851 & 2.954 & 3.725 \\
 & 3  & 3.220 & 4.000 & 52.689 & 85.271 & 0.0461 & 0.0578 & 3.033 & 5.861 & 3.028 & 3.817 \\
 & 5  & 3.186 & 3.953 & 52.759 & 85.305 & 0.0478 & 0.0601 & 2.964 & 5.854 & 3.153 & 3.987 \\
 & 8  & 3.283 & 4.090 & 52.800 & 85.307 & 0.0505 & 0.0636 & 2.941 & 5.912 & 3.180 & 4.033 \\
 & 10 & 3.247 & 4.059 & 52.363 & 85.135 & 0.0509 & 0.0639 & 2.905 & 5.888 & 3.269 & 4.140 \\
\bottomrule
\end{tabular}
\caption{Comparison of SDDP models on datasets with different numbers of factors ($K^*$). Reported values are average MAE and RMSE over 100 runs.}
\label{tab:Sensitive_K}
\end{table*}

\begin{table*}[htbp]
\centering

\begin{tabular}{l|c|cc|cc|cc|cc}
\toprule
\multicolumn{2}{c}{} & \multicolumn{8}{c}{Dataset} \\
\cmidrule(lr){3-10}
\multicolumn{1}{c}{Model} & \multicolumn{1}{c}{$q_{\text{length}}$} &
\multicolumn{2}{c}{Climate} &
\multicolumn{2}{c}{Energy} &
\multicolumn{2}{c}{Light} &
\multicolumn{2}{c}{Weather} \\
\cmidrule(lr){3-4}
\cmidrule(lr){5-6}
\cmidrule(lr){7-8}
\cmidrule(lr){9-10}
& &
MAE & RMSE &
MAE & RMSE &
MAE & RMSE &
MAE & RMSE \\
\midrule
\multirow{4}{*}{SDDP-TCN}
& 50  & 2.972 & 3.710 & 41.843 & 73.098 & 2.851 & 4.924 & 2.669 & 3.357 \\
& 100 & 2.942 & 3.677 & 42.160 & 73.217 & 2.821 & 4.910 & 2.677 & 3.365 \\
& 150 & -- & -- & 42.300 & 73.671 & 2.823 & 4.924 & 2.684 & 3.380 \\
& 200 & -- & -- & 41.231 & 74.683 & 2.790 & 4.936 & 2.680 & 3.371 \\
\midrule
\multirow{4}{*}{SDDP-LSTM}
& 50  & 2.976 & 3.667 & 42.327 & 72.276 & 2.811 & 5.609 & 2.775 & 3.494 \\
& 100 & 2.957 & 3.644 & 42.528 & 72.149 & 2.733 & 5.579 & 2.677 & 3.358 \\
& 150 & -- & -- & 43.954 & 73.381 & 2.756 & 5.561 & 2.710 & 3.404 \\
& 200 & -- & -- & 41.950 & 74.389 & 2.783 & 5.612 & 2.707 & 3.396 \\
\midrule
\multirow{4}{*}{SDDP-DeepAR}
& 50  & 3.187 & 3.927 & 52.673 & 84.207 & 3.039 & 5.938 & 3.088 & 3.889 \\
& 100 & 3.203 & 3.972 & 53.036 & 84.530 & 2.956 & 5.890 & 2.996 & 3.770 \\
& 150 & -- & -- & 53.403 & 84.824 & 2.912 & 5.855 & 3.015 & 3.802 \\
& 200 & -- & -- & 52.804 & 85.306 & 2.885 & 5.857 & 3.046 & 3.831 \\
\midrule
\end{tabular}

\caption{Comparison of window length ($q$) effects across methods and datasets, evaluated by MAE and RMSE. Each experiment is repeated 100 times, and average results are reported. (For the Climate dataset, $q$ is set to 50 and 100, as larger values would exceed roughly one-twentieth of the total sample size. The FinC dataset is excluded due to its short time horizon ($T/20 < 50$), which is incompatible with the chosen window settings.}
\label{tab:sddp_results}
\label{tab:Sensitive_p}
\end{table*}

\subsection{Ablation Study \label{append:ablation_study}}

We further conducted an ablation study to evaluate the contribution of each component in the proposed SDDP framework.
The core procedure of SDDP comprises two primary steps:
(i) generating target-aware predictors $\hat{X}^\star$ using a Deep Neural Network (DNN) based on the Temporal Convolutional Network (TCN), and
(ii) extracting supervised factors $\hat G^\star$ through Principal Component Analysis (PCA).

In addition to the baseline (vanilla) results reported in Table~\ref{tab:Performance comparison},
we include a new variant that removes step~(ii), i.e., omitting PCA and directly using the target-aware predictors for forecasting.

As shown in Table~\ref{tab:ablation_study}, removing either step--(i) the generation of target-aware predictors $\hat{X}^*$ or (ii) the PCA-based dimensionality reduction--leads to a noticeable decline in forecasting performance.
When only step~(i) is retained and PCA is omitted, the average performance drop ranges from approximately 5\% to 15\%, with the Energy dataset being a notable exception.
Conversely, when only step~(ii) is retained and the target-aware transformation is skipped (i.e., applying PCA directly to the raw data), the degradation is more pronounced, averaging 10\%-25\% across datasets.
These findings demonstrate that both components are indispensable: only by jointly applying the target-aware transformation and PCA can the SDDP framework extract highly target-relevant latent factors and achieve optimal predictive accuracy.

\begin{table*}[htbp]
\centering
\small
\setlength{\tabcolsep}{4pt}
\renewcommand{\arraystretch}{0.95}
\begin{tabular}{c|cc|cc|cc|cc|cc}
\toprule
\multirow{2}{*}{\textbf{Model}} 
& \multicolumn{2}{c}{\textbf{Climate}} 
& \multicolumn{2}{c}{\textbf{Energy}} 
& \multicolumn{2}{c}{\textbf{FinC}} 
& \multicolumn{2}{c}{\textbf{Light}} 
& \multicolumn{2}{c}{\textbf{Weather}} \\
\cmidrule(lr){2-3} \cmidrule(lr){4-5} \cmidrule(lr){6-7} \cmidrule(lr){8-9} \cmidrule(lr){10-11}
& MAE & RMSE & MAE & RMSE & MAE & RMSE & MAE & RMSE & MAE & RMSE \\
\midrule
SDDP–TCN           & 2.936 & 3.673 & 42.441 & 74.924 & 0.0446 & 0.0565 & 2.725 & 4.965 & 2.679 & 3.369 \\
Without\_DR–TCN    & 3.076 & 3.814 & 37.814 & 73.515 & 0.0490 & 0.0620 & 4.538 & 6.141 & 2.695 & 3.399 \\
PCA–TCN            & 3.569 & 4.430 & 59.604 & 93.505 & 0.0484 & 0.0653 & 3.555 & 6.344 & 3.832 & 4.782 \\
Vanilla–TCN        & 3.956 & 4.838 & 62.362 & 95.822 & 0.0484 & 0.0654 & 3.494 & 6.333 & 3.991 & 4.974 \\
\midrule
SDDP–LSTM          & 2.946 & 3.628 & 42.107 & 74.451 & 0.0476 & 0.0598 & 2.721 & 5.577 & 2.689 & 3.375 \\
Without\_DR–LSTM   & 3.591 & 4.458 & 39.773 & 71.872 & 0.0481 & 0.0611 & 4.891 & 7.470 & 2.900 & 3.655 \\
PCA–LSTM           & 4.494 & 5.535 & 59.803 & 93.170 & 0.0484 & 0.0654 & 3.583 & 6.504 & 4.434 & 5.509 \\
Vanilla–LSTM       & 4.518 & 5.547 & 59.716 & 93.295 & 0.0484 & 0.0654 & 3.538 & 6.464 & 4.472 & 5.544 \\
\midrule
SDDP–DeepAR        & 3.188 & 3.953 & 52.697 & 85.275 & 0.0476 & 0.0599 & 2.912 & 5.888 & 3.038 & 3.832 \\
Without\_DR–DeepAR & 4.300 & 5.372 & 47.316 & 80.410 & 0.0641 & 0.0798 & 5.395 & 8.147 & 3.333 & 4.196 \\
PCA–DeepAR         & 4.513 & 5.570 & 59.680 & 92.882 & 0.0484 & 0.0653 & 3.629 & 6.513 & 4.445 & 5.526 \\
Vanilla–DeepAR     & 4.532 & 5.575 & 58.979 & 92.850 & 0.0493 & 0.0661 & 3.548 & 6.459 & 4.487 & 5.565 \\
\bottomrule
\end{tabular}
\caption{Ablation study comparing {SDDP}, {Without\_DR} (step (i) only), {PCA} (step (ii) only), and {Vanilla} methods across different DNN architectures and datasets. Each experiment is repeated 100 times, and average MAE and RMSE values are reported.}
\label{tab:ablation_study}
\end{table*}

\section{Proof of Proposition \ref{prop:nonlinear}}

Without loss of generality, assume that
\begin{align}\label{eq:ident}
\frac{1}{T}\sum_{t=1}^T \bx_t \bx_t^\top =I, \qquad \frac1T \sum_t \bg_t^\star \bg_t^{\star\top}=I.    
\end{align}
Let $\bbeta=(\bbeta_1^\top,...,\bbeta_q^\top)^\top$. Based on the linear structure given in Assumption \ref{asmp:linear}, the first step of constructing target-aware predictors is equivalent to fitting a least square estimator. That is,
\begin{align*}
y_{t+h}\approx \widehat\gamma_{i,1} x_{i,t} +\widehat \gamma_{i.2} x_{i,t-1} + \cdots \widehat \gamma_{i,q} x_{i,t-q+1} .    
\end{align*}
Denote $\widehat\bgamma_i=(\widehat\gamma_{i,1},...,\widehat\gamma_{i,q})^\top$ and $\bxi_{i,t}=(u_{i,t},...,u_{i,t-q+1})^\top$. 
By least-squares estimation and the identification condition in \eqref{eq:ident}, we have 
\begin{align}
\begin{aligned}
    \widehat \bgamma_i =  &  \left\{ (\mathbf{I}_q \otimes b_i') \bbeta \right\} + 
\left\{ (\mathbf{I}_q \otimes b_i') \frac{1}{T} \sum_{t=q}^{T-h} \bg_t^\star \epsilon_{t+h} + \right. \\
& \left. \frac{1}{T} \sum_{t=q}^{T-h} \bxi_{i,t} \bg_t^{\star\top} \bbeta + 
\frac{1}{T} \sum_{t=q}^{T-h} \bxi_{i,t} \epsilon_{t+h} \right\} \\ 
:= & \ \bgamma_i + \delta_i.
\end{aligned}
\end{align}

Roughly speaking, each term in $\delta_i$ is of order $O_p(1/\sqrt{T})$. Letting
\begin{align}
\bz_{i,t} = \widehat\gamma_i' (\mathbf{I}_q \otimes b_i') \bg_t^{\star} + \widehat\bgamma_i' \bxi_{i,t},
\end{align}
we have the following lemma.

\begin{lemma} \label{lemma1}
Let $\bZ_t = (\bz_{1,t}, \ldots, \bz_{N,t})'$ and $\widetilde{\mathbf{V}}$ be the diagonal matrix consisting of the top $qK$ eigenvalues of 
\[
\sum_{t=q}^T \bZ_t \bZ_t^\top
\] 
as its diagonal elements. Under Assumption \ref{asmp:linear}, if $N^{1 - \nu} / T^2 \to 0$, with probability tending to one, we have
\[
\widetilde{\mathbf{V}} \asymp N^\nu T.
\]    
\end{lemma}
The proof of Lemma \ref{lemma1} is similar to Lemma 2 in \cite{gao2024supervised}, thus is omitted.

Following similar steps as in the proof of Theorem 1 in \cite{gao2024supervised}, we can derive the desired results. It is worth noting that, unlike \cite{gao2024supervised} and \cite{bai2002determining}, we estimate the factor loading matrix using the top eigenvectors of $\sum_t \widehat\bx_t^\star \widehat\bx_t^{\star\top}$. Nonetheless, the overall proof strategy remains similar.





\end{document}